\documentclass[acmsmall,screen]{acmart}
\AtBeginDocument{%
  }

\setcopyright{acmlicensed}
\copyrightyear{2026}
\acmYear{2026}
\acmDOI{XXXXXXX.XXXXXXX}

\acmJournal{taas}

\usepackage[nameinlink]{cleveref}
\usepackage{tcolorbox}
\usepackage{soul}
\usepackage{multirow}
\usepackage{listings}
\usepackage{algorithmic}
\usepackage{algorithm}
\crefname{figure}{Fig.}{Figs.}
\Crefname{figure}{Fig.}{Figs.}
\crefname{table}{Table}{Tables}
\Crefname{table}{Table}{Tables}
\crefname{algorithm}{Algorithm}{Algorithms}
\Crefname{algorithm}{Algorithm}{Algorithms}
\crefname{equation}{Eq.}{Eqs.}
\Crefname{equation}{Eq.}{Eqs.}
\usepackage{tabularx}
\usepackage{subfig}
\usepackage{makecell}
\usepackage{pifont}
\newcommand{\cmark}{\ding{51}}
\newcommand{\xmark}{\ding{55}}

\begin{document}

\title[Sim2Schedule For Open-Pit Mine Scheduling]{Sim2Schedule: A Simulator-Guided LLM Framework for Autonomous Open-Pit Mine Scheduling}

\author{Mustavi Ibne Masum}
\authornote{Corresponding Author}
\email{mmasum@lakeheadu.ca}
\orcid{0009-0000-7665-4009}
\author{Thiago Eustaquio Alves de Oliveira}
\email{ talvesd@lakeheadu.ca}
\orcid{0000-0002-7164-9064}
\affiliation{%
  \institution{Department of Computer Science, Lakehead University}
  \city{Thunder Bay}
  \state{Ontario}
  \country{Canada}
}

\author{Mahzabeen Emu}
\orcid{0000-0002-0433-1873}
\email{memu@mun.ca}
\affiliation{%
  \institution{Quantum Communications and Computing Research Center and Department of Electrical and Computer Engineering, Memorial University of Newfoundland}
  \city{St. John's}
  \state{Newfoundland}
  \country{Canada}}

\renewcommand{\shortauthors}{Mustavi et al.}

\begin{abstract}
  Open-pit mine scheduling is a critical process for maximizing economic return under complex geotechnical and operational constraints. While Mixed-Integer Linear Programming (MILP) provides mathematically optimal baselines, its exponential computational complexity and inability to adapt in real time limit its practical deployment in dynamic industrial environments. This work introduces a simulator-driven Large Language Model (LLM) scheduling framework in which the LLM acts as an autonomous decision-making agent, guided at each step by a custom simulator that encodes geotechnical precedence, extraction-processing coupling, and dynamic capacity constraints directly into the action generation mechanism. Operating entirely zero-shot within a closed, data-secure environment, the framework produces complete, interpretable extraction and processing schedules without cloud-based inference, domain-specific fine-tuning, or retraining. To provide a trustworthy performance benchmark,  a novel MILP formulation is developed that incorporates realistic operational and geotechnical constraints. Evaluated across mining instances of varying scale and time periods, the LLM-based framework recovers between 94\% and 99\% of the  MILP optimal NPV while scaling linearly in computation time. These results position simulator-constrained LLM agents as a practical and scalable alternative to classical optimization for long-horizon industrial scheduling under complex operational constraints.
\end{abstract}





\maketitle

\section{Introduction}
The extraction of near-surface ore deposits through open-pit mining is a complex industrial process. It involves systematically removing horizontal benches to access valuable minerals. This method requires a careful and ongoing balance between selectively extracting ore for processing and managing large amounts of waste material, or overburden, which must be removed to ensure slope stability and access deeper reserves \cite{Altiti21}. The economics of these operations hinge on a two-part structure: extraction activities, which constitute the primary operational costs, and processing activities, which refine raw ore into marketable products and generate revenue \cite{amponsah}. Determining the most profitable sequence for these activities is known as open-pit production scheduling (OPPS). It is a critical factor in ensuring the long-term economic sustainability of capital-intensive mining projects \cite{cite2}. The central objective is to maximize the net present value (NPV) of the operation by timing each block's extraction optimally, navigating a fundamental economic trade-off where excavating overburden incurs immediate costs yet remains a mandatory prerequisite for exposing high-grade ore at depth~\cite{saavedra2016optimizing}. The spatial and temporal optimization of this sequence is thus the primary challenge in OPPS.

As Large Language Models (LLMs) have become a revolutionary approach in artificial intelligence, their ability to make interpretable decisions and engage in contextual reasoning has drawn interest in addition to natural language tasks~\cite{li2024pre}. In mine planning specifically, recent works have explored data-driven and AI-assisted approaches to operational scheduling, including short-horizon truck dispatching and task sequencing \cite{icarte2025intelligent}, yet the application of LLMs to the long-horizon strategic problem of block extraction and sequencing remains unexplored. This gap is significant, as LLMs offer real-time adaptability and zero-shot generalization that could substantially reduce the burden of expert re-optimization when operational conditions change \cite{bubeck2023sparks, da2025large, yang2023large}. 

Deploying LLMs for mine scheduling introduces practical challenges. LLMs demand considerable computational and storage resources during training and inference \cite{wang2025parameter}, which is a substantial burden in resource-constrained industrial environments. If the LLM is used purely for inference, without any domain-specific fine-tuning, the computational overhead is dramatically reduced. This observation motivates the central contribution of this work, which introduces a framework in which an LLM acts solely as an inference-time decision-maker, guided at each step by a custom simulator that encodes domain knowledge directly.

\begin{figure}
    \centering
    \includegraphics[width=0.9\linewidth]{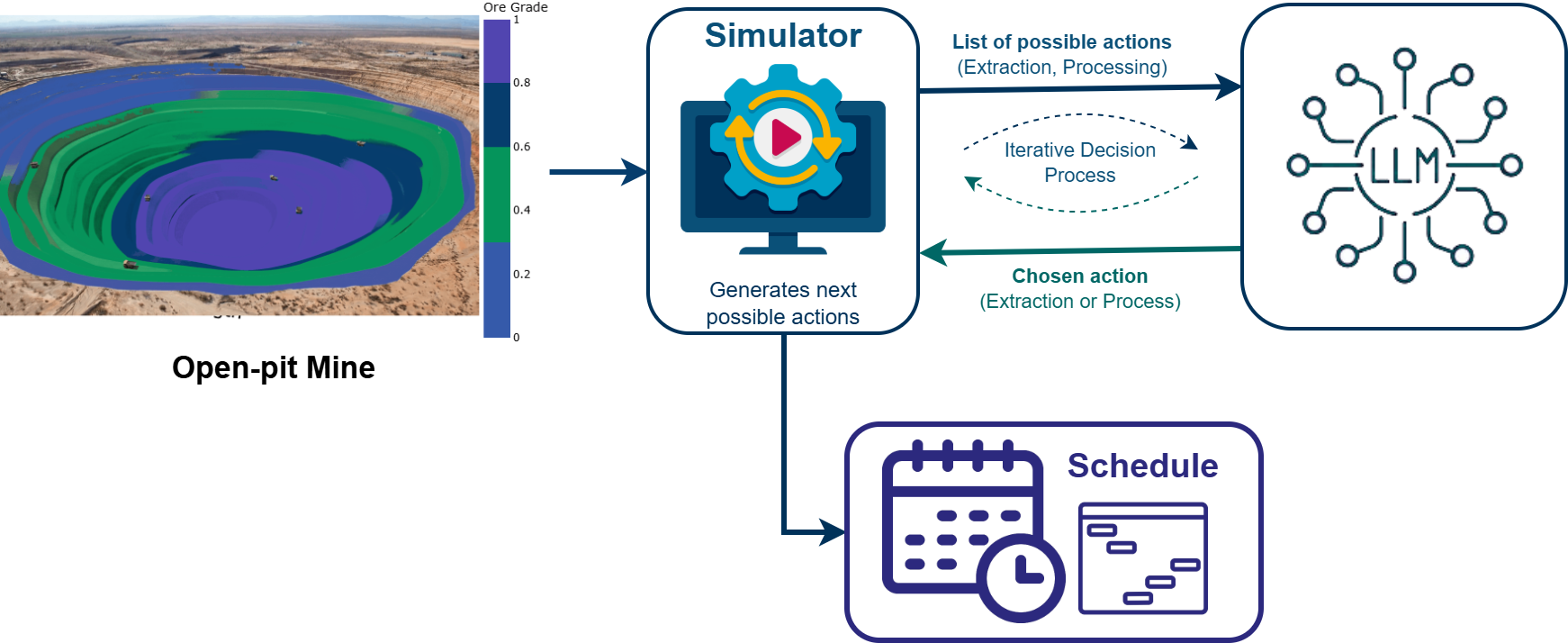}
    \caption{Overview of the LLM-Assisted Scheduling Framework}
    \label{fig:methodsimple}
    \Description{Overview diagram of a proposed LLM-based simulation framework for open-pit mine extraction scheduling.}
\end{figure}

As illustrated in~\cref{fig:methodsimple}, the simulator functions as the medium through which mine-state knowledge is communicated to the LLM and through which decisions are received and recorded. At each scheduling step, the simulator evaluates the current excavation state and identifies all physically feasible actions. These actions are then presented to the LLM as structured context. Once the LLM selects an action, the simulator processes the decision and advances the mine state accordingly. This results in a complete extraction and processing schedule generated entirely without any LLM training, fine-tuning, or adaptation. The framework operates as a closed, self-contained system with no cloud-based LLM calls, ensuring that sensitive mine data remains within the operational environment.

Given the computational complexity of OPPS, Mixed-Integer Linear Programming (MILP) has historically served as the dominant optimization paradigm, offering formal integration of economic, technical, and operational constraints. However, existing MILP formulations frequently ignore the necessity of safely removing overlying rock before deeper extraction, permit the mathematical processing of partially buried material, and impose rigid production capacity targets that become infeasible as the mine depletes. To address these deficiencies, this work introduces a novel MILP formulation that explicitly enforces these physical realities, yielding genuinely executable schedules. While this improves operational realism, the resulting formulations are large-scale and computationally expensive to solve, require expert reformulation whenever new operational variabilities arise, and produce sequences that demand post-processing before they are interpretable by mine planners.

These limitations motivate the need for scheduling frameworks that balance mathematical rigor with real-time adaptability and interpretability. While metaheuristics and reinforcement learning have been explored for this purpose, they frequently struggle with generalizability and transparent reasoning across diverse mining contexts \cite{app151810033}. The LLM-driven simulator framework proposed here addresses these shortcomings by providing a sequential, interpretable decision process that adapts to operational changes without re-optimization. The MILP formulation serves in this study not as the proposed solution, but as the optimal benchmark against which the LLM agent is evaluated alongside a Greedy heuristic and a Random baseline, enabling a systematic assessment of whether LLM-driven scheduling can effectively navigate the trade-off between computational tractability and solution quality.
The main contributions of this study are as follows:
\begin{itemize}

     \item \textbf{LLM-Based Mining Agent for Long-Term OPPS}: An autonomous scheduling framework is designed in which an LLM functions as the decision-making agent. Unlike static solvers, this agent generates {extraction processing} sequences dynamically by evaluating the evolving set of feasible actions at each time step. Unlike prior mining AI, which is restricted to short-horizon operational tasks such as truck dispatching~\cite{icarte2025intelligent}, our agent addresses the multi-period, strategic problem of block extraction and processing sequencing for NPV maximization. The LLM operates zero-shot within a custom simulator, requiring no domain-specific training and producing immediately interpretable, executable schedules.

    \item \textbf{Interpretable, Open-Source Mining Simulator}: An interactive open-pit scheduling simulator is developed, which, to the best of our knowledge, is the first of its kind to be integrated with LLM agents. The simulator assesses viable actions, processes decisions, and updates the system state in real-time. Crucially, it generates decision sequences that are transparent and immediately interpretable by mine planners, facilitating operational validation. The simulator is made publicly available to support reproducibility and further methodological development.\footnote{The simulation tool is available at:
    {https://anonymous.4open.science/r/LLM-MILP-Mining-DF4E}}
    
    \item \textbf{MILP Formulation for Realistic Constraints}: 
    A robust MILP model is developed to resolve critical oversimplifications in traditional scheduling. Previous formulations rely on rigid capacity limits and physically impossible partial processing, producing seemingly optimal but geotechnically invalid schedules. Our approach integrates dynamic capacity bounds, strict spatial precedence for safety, and exact extraction processing coupling. By enforcing these realistic constraints alongside standard production and grade blending requirements, the formulation generates optimal NPV maximizing plans that are mathematically sound and genuinely executable.

\end{itemize}
The remainder of this paper is organized as follows:~\cref{sec:lit} reviews the relevant literature, ~\cref{sec:framework} introduces our proposed framework and methodology, and~\cref{sec:result} details the experimental setup, data configuration, and comparative evaluation results. Finally, the \cref{sec:conclusion} summarizes the key findings and concludes the study.
\section{Literature Review}\label{sec:lit}
OPPS represents one of the most challenging optimization problems in mine planning, defined as the process of determining the optimal time and sequence for extracting blocks from a mine to maximize NPV while satisfying physical, environmental, operational, and economic constraints \cite{caccetta2003application}.
\subsection{Mathematical Programming Approaches}
Early foundations were laid by Lerchs \textit{et al.}~\cite{lerchs1965}, who revolutionized ultimate pit definition using graph-theoretic methods, and by Johnson~\cite{johnson}, who established the analytical basis for OPPS using Linear Programming (LP). However, early LP models frequently yielded fractional block extractions, requiring heuristic post-processing to produce executable schedules. The transition to MILP addressed this by enforcing discrete decision variables, establishing it as the dominant paradigm. Askari-Nasab \textit{et al.}~\cite{askari} advanced this direction through clustering and mining-cut aggregation, improving both computational tractability and NPV. Despite these improvements, standard MILP formulations often rely on relaxed precedence structures and static extraction limits to remain solvable, reducing their operational realism.

Subsequent work attempted to address specific operational constraints. Khazaei \textit{et al.}~\cite{Khazaei} incorporated vertical and horizontal precedence for Sub-Level Caving, though their formulation lacks strict extraction-processing coupling. Rahnema \textit{et al.}~\cite{rahnema} integrated GHG emissions and carbon-tax penalties into a deterministic MILP, demonstrating that environmental factors can be incorporated without altering the core structure, though simplified sequencing logic and fixed thresholds limit dynamic consistency. Tabesh \textit{et al.}~\cite{Tabesh} introduced a two-stage clustering framework that improves tractability and stabilizes feed quality yet retains non-adaptive constraints that do not respond to temporal changes in mine state. Stochastic extensions have also been explored to address grade uncertainty, with simultaneous optimization of entire mining complexes demonstrating NPV improvements under geological variability~\cite{Dimitrakopoulos03072022}, though these retain the same computational scalability limitations as deterministic models. To overcome the intractability of exact solvers on large-scale instances, metaheuristic alternatives have been explored. Amponsah \textit{et al.}~\cite{amponsah} compared a genetic algorithm framework against a stochastic MILP under grade uncertainty, showing significant runtime reductions at the cost of strict optimality. Hussain \textit{et al.}~\cite{hussain} demonstrated, outside the mining domain, that MILP can handle complex multi-period physics and strict constraint coupling, though equivalent dynamic consistency remains elusive in mine planning. Collectively, these methods share a fundamental limitation, functioning as static, batch computational engines that require expert reformulation whenever operational conditions change.

\subsection{LLM-Based and Simulation-Driven Approaches}
LLMs have recently emerged as powerful engines for combinatorial optimization, offering a qualitatively different approach from both exact solvers and metaheuristics. Recent work has demonstrated that LLMs can serve as end-to-end solvers for NP-hard combinatorial problems, with feasibility-aware models achieving optimality gaps as low as 1.03\% on benchmark instances \cite{jiang2025large}. The StarJob framework~\cite{starjob2024} validated this potential in Job Shop Scheduling, where fine-tuned LLMs outperformed traditional dispatching rules. In logistics, Agentic Frameworks with LLMs~\cite{afl2025} have been applied to Vehicle Routing Problems, where LLM agents interact with external solvers to enforce constraints that pure language models often violate. Multi-agent LLM frameworks have further demonstrated that feedback-driven iterative loops between agents can adaptively refine scheduling solutions across diverse problem types without algorithm redesign~\cite{maaef2025}.

A particularly compelling direction is the pairing of LLMs with dedicated simulators or physics-enforcing environments. Ma \textit{et al.}~\cite{ma2024bilevel} formalized this as a bilevel optimization framework, showing that LLMs optimizing discrete scientific hypotheses, when coupled with a physical simulator handling continuous parameter evaluation, can outperform methods relying on either component alone. Similarly, work on LLM-empowered agent-based modeling and simulation has shown that offloading constraint enforcement and numerical computation to a structured environment, while reserving high-level reasoning for the LLM, significantly reduces hallucination and improves decision quality in complex dynamic settings~\cite{gao2024llm}. This simulator-in-the-loop paradigm is directly relevant to OPPS, where the mine state evolves continuously and each decision constrains all future options in a cascading, non-reversible manner that demands both physical consistency and forward-looking reasoning.

Despite this progress, a critical gap remains. Existing mining-related AI models primarily address short-term truck dispatching and simulation-based operational planning~\cite{icarte2025intelligent, UPADHYAY2018153}, and no prior LLM-based scheduling work contends with the 3D geotechnical precedence, extraction-processing coupling, or multi-period NPV objectives that define OPPS. LLM-based scheduling frameworks such as A4PS~\cite{LI2026207} have shown promise in smart manufacturing contexts but address job-sequencing in flexible production cells rather than the spatially constrained, long-horizon problem of open-pit mining. Furthermore, LLMs pre-trained on generic data struggle with complex physical constraints without external knowledge structures~\cite{ieee}, and Chain-of-Thought prompting alone~\cite{wei2022chain} is insufficient to enforce the hard feasibility requirements of mine scheduling. 

As summarized in~\cref{tab:milp_comparison}, no prior method simultaneously enforces strict precedence, extraction-processing coupling, and dynamic capacity limits while remaining accessible to mine planners~\cite{amponsah, askari, hussain, Khazaei, rahnema, Tabesh}. The proposed methods address this gap from complementary directions, with the MILP providing an optimal but computationally intensive benchmark and the LLM-based simulator matching its constraint coverage while delivering superior scalability and real-time planner transparency.

\begin{table}
\centering
\caption{Comparison of deterministic and metaheuristic optimization approaches for mine production scheduling}
\label{tab:milp_comparison}
\resizebox{\columnwidth}{!}{%
\begin{tabular}{l| c c| c c c |c} 
\hline 
& \multicolumn{2}{|c}{\textbf{Model Performance}} & \multicolumn{3}{|c|}{\textbf{Constraints}} & \\
\cmidrule{2-3} 
\cmidrule{4-6}
\textbf{Reference} & 
\textbf{Optimality} & \textbf{Scalability} & 
\makecell{\textbf{Strict}\\\textbf{Precedence}} & 
\makecell{\textbf{Extraction-Processing}\\\textbf{Coupling}} & 
\makecell{\textbf{Dynamic}\\\textbf{Limits}} & 
\makecell{\textbf{Accessibility to}\\\textbf{Mine Planners}} \\ 
\midrule 
Johnson~\cite{johnson} & \xmark & \xmark & \xmark & \xmark & \xmark & \xmark \\ 

Askari-Nasab \textit{et al.}~\cite{askari} & \cmark & \xmark & \xmark & \xmark & \xmark & \xmark \\ 

Khazaei \textit{et al.}~\cite{Khazaei} & \cmark & \xmark & \xmark & \xmark & \xmark & \xmark \\ 

Rahnema \textit{et al.}~\cite{rahnema} & \cmark & \xmark & \xmark & \xmark & \xmark & \xmark \\ 

Tabesh \textit{et al.}~\cite{Tabesh} & \cmark & \cmark & \xmark & \xmark & \xmark & \xmark \\ 

Hussain \textit{et al.}~\cite{hussain} & \cmark & \xmark & \cmark & \cmark & \xmark & \xmark \\ 

 {Amponsah \textit{et al.}~\cite{amponsah}} &  {\xmark} &  {\cmark} &  {\cmark} &  {\xmark} &  {\xmark} &  {\xmark} \\

 {\textbf{Proposed MILP Method}} &  {\textbf{\cmark}} &  {\textbf{\xmark}} &  {\textbf{\cmark}} &  {\textbf{\cmark}} &  {\textbf{\cmark}} &  {\textbf{\xmark}} \\

 {\textbf{Proposed LLM Method}} &  {\textbf{\xmark}} &  {\textbf{\cmark}} &  {\textbf{\cmark}} &  {\textbf{\cmark}} &  {\textbf{\cmark}} &  {\textbf{\cmark}} \\
\bottomrule 
\end{tabular} 
}
\end{table}

\section{Methodology}\label{sec:framework}
This section presents the comprehensive framework developed to address the OPPS problem through multiple solution approaches. It begins by formulating the general OPPS problem and defining the operational constraints that govern feasible mining sequences. The proposed dual-pathway framework is then described, centering on the iterative simulator and LLM-driven decision-making pipeline, with the MILP model serving as the optimal benchmark for comparative evaluation. Finally, the simulator-based heuristic approaches are described in detail, followed by the mathematical formulation of the deterministic MILP model.

\subsection{Problem Formulation}
The OPPS problem seeks to maximize the overall profitability of a mining operation by optimally sequencing block extraction and ore processing across a finite time period. The fundamental task is to determine how much ore and waste material to mine from each block in each period, how much ore to process, and when each block becomes available for extraction based on slope stability requirements.
The economic objective is formalized by maximizing the project's NPV, which evaluates overall profitability by summing the net cash flows, defined as the periodic revenues from processed ore minus material extraction costs, divided by a discount factor raised to the power of the respective time period~\cite{amponsah}.
Achieving this objective requires satisfying multiple categories of operational and physical constraints that collectively ensure feasibility and economic viability.

Capacity constraints enforce upper and lower bounds on total material extraction per period based on equipment limitations. However, in practical mining operations, production cannot always adhere strictly to fixed minimum or maximum capacity limits, particularly near the end of the mine's useful life. While existing MILP formulations improve NPV and computational tractability through various approaches \cite{amponsah, Tabesh}, these models rely on statically fixed capacity limits that do not respond to temporal changes or diminishing reserves in the mine state. To address this limitation, the proposed model incorporates flexible capacity constraints through a dynamic lower bound mechanism that adjusts minimum extraction levels based on remaining reserves, preventing mathematical infeasibility while maintaining realistic production behavior. Processing capacity constraints bound the throughput of ore treated in each period within the mill's operational range, while quality constraints ensure that the weighted average grade of ore processed in each period satisfies upper and lower specifications to maintain optimal metallurgical performance through grade blending requirements.

\begin{figure}
\centering
\includegraphics[width=0.5\linewidth]{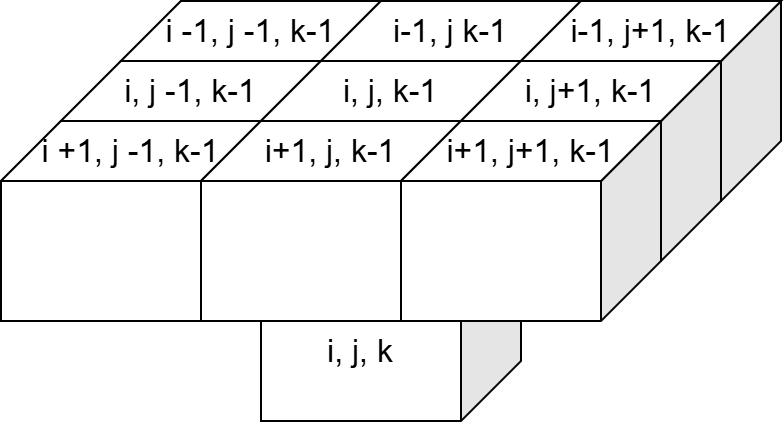}
\caption{Precedence relationships in a 3D block-structured grid. A block located at $(i,j,k)$ can be extracted only after the overlying blocks $(i-1,j-1,k-1), \ldots, (i+1,j+1,k-1)$ have been completely removed.}
\label{fig:precedence}
\Description{3D grid of stacked blocks representing a mine model. A single target block at position (i, j, k) is highlighted, with arrows or shading indicating the cone-shaped group of overlying blocks in the layer above (k-1) that must be removed before the target block can be extracted.}
\end{figure}

Geotechnical constraints enforce block precedence relationships that dictate the physical sequence of extraction. As illustrated in~\cref{fig:precedence}, for a block located at coordinates $(i,j,k)$ to be accessible, the overlying blocks in the range $(i-1,j-1,k-1),\ldots,(i+1,j+1,k-1)$ must first be completely removed. This geometric dependency ensures slope stability and operational feasibility by preventing the extraction of blocks before their supporting overburden has been cleared. The precedence constraint explicitly defines the set of predecessor blocks for each block in the orebody, creating a complex network of dependencies that must be satisfied throughout the extraction sequence.
Another critical aspect of the formulation is the strict coupling between extraction and processing. Existing formulations~\cite{amponsah,askari} do not enforce the requirement that a block must be completely extracted before its ore can be routed to the processing plant, allowing partial extraction and immediate processing in the same period. This contradicts real-world mining physics where a block must be fully blasted and excavated before its constituent ore can be separated and sent to the mill. To address these deficiencies, the proposed model incorporates logical and coupling constraints that ensure each block is mined at most once and processed at most once across all periods, while enforcing the extraction-processing dependency that ore from a block can only be processed after the block has been fully extracted. Temporal consistency requirements further ensure that once a block becomes available for extraction, it remains available in all subsequent periods.
\subsection{Proposed Framework}
\begin{figure}
\centering
\includegraphics[width=\linewidth]{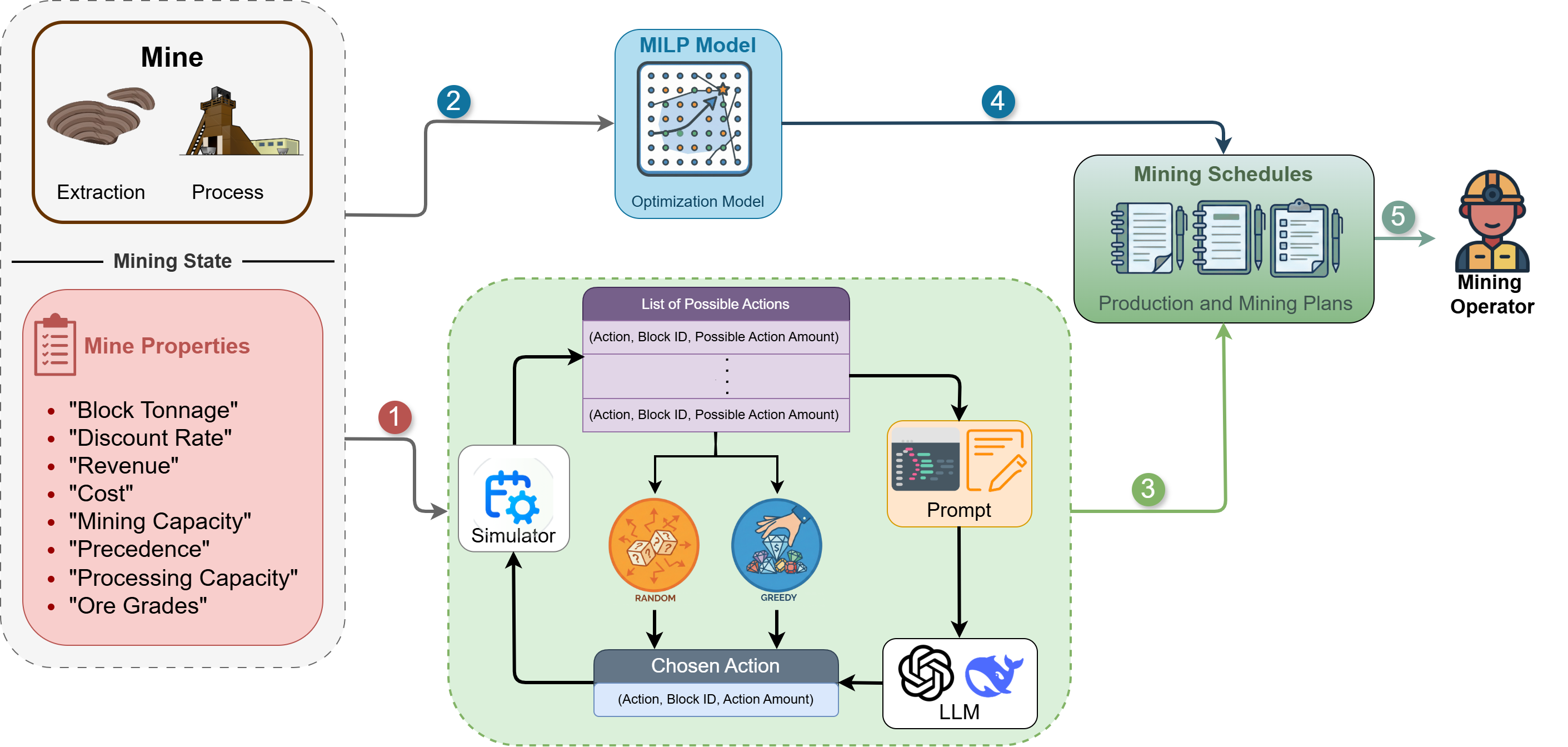}
\caption{Overview of the Proposed Framework. Starting from the initial mining state, the workflow splits into two parallel evaluation paths. Path 1 feeds the mine properties into an iterative simulator where Random, Greedy, and LLM agents generate heuristic schedules (3). Path 2 passes the same data through a MILP optimization model to produce an optimal deterministic schedule (4). Both outputs are aggregated in the Mining Schedules module and presented to the mining operator (5) for comparative analysis.}
\label{fig:methodology}
\Description{Flowchart of a proposed mine scheduling framework. Starting from an initial mining state, the workflow branches into two parallel paths. The primary path feeds into an iterative simulator containing three agents, Random, Greedy, and LLM, which generate heuristic schedules. The secondary path passes through a MILP model to produce an optimal deterministic schedule. Both paths converge where aggregated results are presented to the mining operator for comparative analysis.}
\end{figure}

Having established the problem formulation and constraint requirements, this section presents the overall computational framework that enables systematic comparison between the LLM-driven simulator and the MILP benchmark. As illustrated in~\cref{fig:methodology}, the workflow originates from a shared mining environment where initial mine properties, including block tonnage, ore grades, and operational capacities, define the system state. This shared input initializes two parallel computational streams, ensuring that all methods operate under identical operational assumptions and enabling fair performance comparisons.

The primary stream (labeled 1) directs the initial mine state into the iterative simulator framework, inspired by the ReAct paradigm~\cite{yao2023react}. The simulator functions as the operational core of the framework, dynamically managing all physical and operational requirements by generating only feasible actions at each decision step. This implicit constraint enforcement covers capacity, quality, precedence, and extraction-processing coupling, entirely through the action generation mechanism rather than through explicit mathematical programming. At each simulation step, the simulator observes the current mine state and produces a structured list of feasible actions, each specifying an action type, block ID, and admissible quantity. Three independent simulation experiments are conducted concurrently within this environment, each employing a distinct decision-making strategy: Random selection, Greedy heuristic, and LLM-based reasoning with structured prompts. Each agent independently selects an action according to its strategy, which updates the state exclusively within its respective simulation branch. This process repeats iteratively until no further feasible actions remain or the terminal time period is reached. The LLM agent receives the current feasible action list alongside contextual mine state information through a structured prompt, and reasons over the long-term implications of each candidate action before committing to a selection. The resulting heuristic schedules are transferred to the Mining Schedules module for aggregation (labeled 3).

The secondary stream (labeled 2) passes the same initial data into the deterministic MILP formulation, which optimizes production scheduling based on explicitly defined economic objectives and operational constraints. These constraints are encoded as mathematical inequalities, with continuous variables modeling fractional block extraction and processing, binary variables enforcing precedence and logical dependencies, and auxiliary variables handling conditional constraints such as dynamic capacity bounds and extraction-processing coupling. The raw solver output is processed by a schedule decoder that translates the optimal mathematical solution into an interpretable deterministic schedule (labeled 4), which is then stored in the Mining Schedules module alongside the simulator outputs.

The aggregated results from both streams are finally presented to the mining operator (labeled 5), enabling direct comparison of the optimal MILP schedule against the three simulator-generated heuristic schedules across NPV, constraint satisfaction, and computational efficiency.
\subsection{Simulator-Based Heuristic Approaches}
Having established the problem formulation and framework, this section describes the primary contribution of this work, a custom-built iterative simulator that serves as the operational environment for heuristic and LLM-driven scheduling agents. Unlike traditional simulation-based approaches that focus on short-term operational planning \cite{UPADHYAY2018153}, the proposed simulator addresses the long-horizon strategic problem of block extraction and processing sequencing. Rather than solving the entire scheduling horizon as a single global optimization as the MILP does, the simulator operates step-by-step, generating and enforcing feasible actions at each discrete decision point, enabling real-time adaptability and transparent decision-making throughout the scheduling process.

\subsubsection{Simulator Architecture and State Management}
Initially, the simulator imports all pertinent mine data, such as block tonnage, mineral grades, revenues, costs, discount rates, processing capacities, and spatial coordinates.
The full operational environment is constructed by establishing the precedence graph, coupling constraints, and capacity boundaries that govern which actions are physically and operationally admissible at any moment in the simulation using this data. Once initialized, the simulator evaluates the current mine structure and generates a list of possible actions available at each step. 
At the start of the simulation, only surface-level blocks with no outstanding predecessors are eligible for extraction. As the simulation progresses and blocks are cleared, the precedence graph is updated and new blocks become accessible, expanding the feasible action set dynamically. This mirrors the real progressive nature of open-pit excavation, where access to deeper material is continuously unlocked as overlying material is removed. The constraint-aware action generation mechanism forms the operational core of the simulator. Rather than requiring agents to independently verify complex operational constraints such as capacity limits, grade blending requirements, precedence dependencies, and extraction-processing coupling, the simulator enforces all of these internally during action generation, presenting only strictly feasible actions to the agent at each step. When an agent selects an action, the simulator validates that selection against the current mine state before committing it to the schedule. If the chosen action is found to be inadmissible under the current operational conditions, the simulator rejects it and requests a new selection from the agent, ensuring that no infeasible decisions are ever recorded regardless of the agent strategy being employed. This separation between constraint enforcement by the simulator and value-based selection by the agent allows each approach to focus entirely on optimization reasoning.
The simulator maintains a complete record of all actions taken, organized by time period. Extraction and processing decisions are logged as they occur, meaning the evolving schedule can be inspected and interpreted in real time as it is being generated. At the conclusion of the simulation, the full schedule is available as a structured record showing exactly which blocks were extracted or processed, in which time period, and by which action type. This transparency makes the simulator output immediately interpretable by mine planners without any post-processing, distinguishing it fundamentally from raw MILP solver output. Schedule NPV is computed using~\cref{obj} to maintain mathematical consistency with the MILP benchmark.
\subsubsection{Baseline Agents}
To establish meaningful performance boundaries, the simulator evaluates agent strategies against two baselines that represent the extremes of decision-making sophistication. A Random agent sets the lower performance threshold by selecting actions uniformly at random from the feasible set at each step, serving as a reference for completely unguided decision-making and providing a floor below which no reasonable strategy should fall. A Greedy agent maximizes instantaneous NPV at each step by always selecting the action with the highest immediate economic return, without any consideration of how that choice affects future accessibility or long-term sequencing. While the Greedy agent performs better than Random in most cases, its short-sighted nature means it can inadvertently delay the removal of waste necessary to unlock high-value ore in later periods, ultimately reducing the discounted NPV of the full schedule. Benchmarking against both baselines provides a calibrated performance range within which the LLM-based reasoning agent is evaluated.
\subsubsection{LLM-Based Simulation Framework}
The LLM agent operates within the iterative action-state loop of the simulator, analogous to a discrete-event control architecture where the simulator functions as the plant model and the LLM functions as the autonomous controller. At each decision step, the LLM receives the current mine state and feasible action list from the simulator and returns a selected action, with no operator input required during execution. This zero-shot operational mode means the LLM relies entirely on its pre-trained reasoning capacity and the structured contextual information provided by the simulator prompt, without any domain-specific fine-tuning or retraining.

At each step, the simulator provides the LLM with the updated system state and the current feasible action list, with each candidate action characterized by its operational type, block identifier, admissible quantity, and marginal economic value. Scheduling begins with extraction actions, where the LLM is presented with block identifiers and their associated unit costs. Despite an inherent tendency in LLMs to avoid actions with immediate negative cash flow, the simulator enforces minimum capacity requirements that oblige the model to commit to extractions in order to maintain mine progression. As surface and near-surface blocks are cleared, processing actions become eligible, creating opportunities for the LLM to reason strategically about extraction-processing sequencing and its implications for discounted long-term NPV. The simulation loop terminates when no viable actions remain in the feasible set or the life-of-mine horizon is reached.
\begin{algorithm}
\caption{LLM-Guided Mining Scheduling}
\label{alg:llm_scheduler}
\begin{algorithmic}[1]
\REQUIRE Set of actions $\mathcal{A}$, max time $T_{max}$, 
processing cap $Qu^t$, extraction cap $Cu^t$, 
System Prompt $\mathcal{P}_{sys}$, Simulator Prompt $\mathcal{P}_{sim}$
\ENSURE Schedule $\mathcal{S}$, total NPV $V_{total}$

\STATE Initialize LLM with System Prompt $\mathcal{P}_{sys}$ \COMMENT{Define role and optimization goals}
\STATE $t \gets 1$
\STATE $V_{total} \gets 0$
\STATE $R_{proc} \gets Qu^1$, $R_{ext} \gets Cu^1$
\STATE $\mathcal{S} \gets \emptyset$

\WHILE{$\mathcal{A} \neq \emptyset$ AND $t \leq T_{max}$}
    \STATE Scheduler computes feasible set $\mathcal{F} \subseteq \mathcal{A}$

    \IF{$\mathcal{F} = \emptyset$}
        \STATE $t \gets t + 1$
        \STATE $R_{proc} \gets Qu^t$, $R_{ext} \gets Cu^t$
        \STATE \textbf{continue}
    \ENDIF

    \STATE Compute $\Delta NPV_i$ for $i \in \mathcal{F}$ \COMMENT{Prepare metrics for context}
    \STATE Construct Simulator Prompt $\mathcal{P}_{sim}$ with candidate actions $\mathcal{F}$
    \STATE $a^* \gets \text{LLM}(\mathcal{P}_{sim})$ \COMMENT{LLM selects action via simulator prompt}

    \STATE $V_{total} \gets V_{total} + \Delta NPV_{a^*}$
    \STATE $\mathcal{S} \gets \mathcal{S} \cup \{(a^*, t)\}$
    \STATE $\mathcal{A} \gets \mathcal{A} \setminus \{a^*\}$

    \STATE $R_{proc} \gets R_{proc} - use_{proc}(a^*)$
    \STATE $R_{ext} \gets R_{ext} - use_{ext}(a^*)$

    \IF{$R_{proc} \le 0$ OR $R_{ext} \le 0$}
        \STATE $t \gets t + 1$
        \STATE $R_{proc} \gets Qu^t$, $R_{ext} \gets Cu^t$
    \ENDIF
\ENDWHILE

\RETURN $\mathcal{S}, V_{total}$
\end{algorithmic}
\end{algorithm}

\cref{alg:llm_scheduler} formalizes this procedure. The process initiates by configuring the LLM with a system prompt $\mathcal{P}{sys}$ that defines its optimization role and establishes the decision-making context, followed by the initialization of state variables, including the time index $t$, cumulative NPV $V{total}$, and the remaining processing ($R_{proc}$) and extraction ($R_{ext}$) capacities (line numbers 1--5). The core scheduling loop continues as long as schedulable actions remain and the maximum time period $T_{max}$ has not been exceeded (line number 6). In each iteration, the scheduler identifies the set of feasible actions $\mathcal{F}$ based on current geotechnical precedence and resource constraints (line number 7). If no actions are feasible in the current state, the scheduler advances to the next time period and resets $R_{proc}$ and $R_{ext}$ to their full capacity limits ($Qu^t$, $Cu^t$) (line numbers 8--12).

When valid candidates exist, the simulator computes the marginal NPV impact $\Delta NPV_i$ for each feasible action and constructs a context-aware prompt $\mathcal{P}{sim}$ that presents these candidate actions to the LLM (line numbers 13--14). The LLM then evaluates this context and selects the action $a^*$ that it determines to be optimal based on the provided information (line number 15). Following this decision, the system updates the global schedule $\mathcal{S}$ by recording the selected action and its execution time, accumulates the realized NPV contribution $\Delta NPV{a^*}$ to the total value $V_{total}$, and removes the executed action from the available set $\mathcal{A}$ (line numbers 16--18). The remaining capacities $R_{proc}$ and $R_{ext}$ are then decremented by the resource usage of the selected action (line numbers 19--20). If either resource is exhausted, the algorithm automatically increments the time period and restores full capacities for the subsequent cycle (line numbers 21--24). This iterative process continues until all actions have been scheduled or the time period is reached, at which point the final schedule $\mathcal{S}$ and total NPV $V_{total}$ are returned.


\subsection{Deterministic MILP Formulation}
A deterministic MILP model is formulated to establish an optimal baseline for evaluating the proposed heuristic and LLM-driven methods.
While Amponsah \textit{et al.}~\cite{amponsah} proposed a foundational MILP formulation for a similar problem, their approach with many traditional models exhibits significant practical drawbacks that prevent it from generating operationally realistic schedules. To produce a benchmark that mine planners can genuinely trust, the proposed model directly resolves these deficiencies across three critical dimensions.

First, traditional fixed minimum extraction thresholds cause mathematical infeasibility near end-of-mine-life when insufficient material remains to satisfy the lower bound. The proposed model addresses this through a dynamic lower bound mechanism that scales the minimum extraction threshold to the actual remaining reserves in each period, allowing the operation to wind down naturally without artificial period extensions or unnecessary costs. Second, many existing formulations simplify the predecessor set in ways that yield physically impossible or geotechnically unsafe extraction sequences. The proposed model explicitly constructs the full 3D predecessor set from slope geometry, encoding realistic spatial precedence requirements directly into the constraint structure. Third, prior formulations permit a block to be partially extracted and simultaneously routed to the processing plant within the same period, which contradicts the physical reality that material must be fully blasted and excavated before it can be separated and milled. Auxiliary binary variables are introduced to enforce strict extraction-processing coupling, mathematically guaranteeing that no ore is processed before its host block is completely mined.

Specifically, the base constraints \cref{equ: c1,equ: c5,equ: c6,equ: c7,equ: c8,equ: c9,equ: c10,equ: c11,equ: c12,equ: c13} are adapted from prior work, while constraints~\cref{equ: c2,equ: c3,equ: c4,equ: c14,equ: c15} represent the novel contributions introduced in this study. The mathematical notation is first defined, followed by the objective function and the complete set of constraints.~\cref{tab:variable} summarizes the indices, sets, parameters, and decision variables used in the formulation. 
\begin{table}[ht]
\caption{Model indices, sets, parameters, and decision variables}
\centering
\begin{tabularx}{\columnwidth}{p{2cm}X}
\toprule
\textbf{Notation} & \textbf{Description} \\
\midrule

\multicolumn{2}{l}{\textit{\textbf{Indices and Sets}}} \\

$n \in \{1,\ldots,N\}$ 
& Index for blocks. \\

$t \in \{1,\ldots,T\}$ 
& Index for scheduling periods. \\

$H_n(D)$ 
& Set of immediate predecessor blocks that must be extracted before block $n$ to satisfy slope constraints; $D$ denotes the total number of predecessor blocks. \\

\midrule
\multicolumn{2}{l}{\textit{\textbf{Parameters}}} \\

$r$ 
& Discount rate. \\

$o_n$ 
& Ore tonnage in block $n$. \\

$w_n$ 
& Waste tonnage in block $n$. \\

$v_n^t$ 
& Revenue from selling the final product of block $n$ in period $t$, minus the discounted mining cost treating all material as ore. \\

$q_n^t$ 
& Cost of mining all material in block $n$ in period $t$ as waste. \\

$re_n^t$ 
& Remaining extractable fraction of block $n$ in period $t$. \\

$Cl^t$ 
& Lower bound of mining capacity in period $t$. \\

$Cu^t$ 
& Upper bound of mining capacity in period $t$. \\

$Ql^t$ 
& Lower bound of processing capacity in period $t$. \\

$Qu^t$ 
& Upper bound of processing capacity in period $t$. \\

$g_n$ 
& Average grade in the ore portion of block $n$. \\

$\underline{g}^t$ 
& Lower bound of required average head grade in period $t$. \\

$\bar{g}^t$ 
& Upper bound of required average head grade in period $t$. \\

$M$ 
& Sufficiently large constant (e.g., $10^4$). \\

\midrule
\multicolumn{2}{l}{\textit{\textbf{Decision Variables}}} \\

$x_n^t \in [0,1]$ 
& Portion of block $n$ extracted as ore and processed in period $t$. \\

$y_n^t \in [0,1]$ 
& Portion of block $n$ mined in period $t$, encompassing both ore and waste fractions. \\

$b_n^t \in \{0,1\}$ 
& Equal to 1 if extraction of block $n$ has started by period $t$; controls extraction precedence. \\

$l^t \in \{0,1\}$ 
& Equal to 1 if remaining material exceeds the lower mining capacity bound in period $t$, activating the minimum limit. \\

$u_n^t \in \{0,1\}$ 
& Equal to 1 if block $n$ has been fully mined by period $t$, enabling its processing. \\

\bottomrule
\end{tabularx}
\label{tab:variable}
\end{table}

The NPV of OPPS is determined by the economic block value of distinct blocks within the orebody block model. The cost of mining a block depends on its depth from the surface and distance to its final destination, and the NPV is computed by discounting these values across all time periods to reflect the time value of money.
\subsubsection{Objective Function}
The primary goal of the MILP model is to maximize the NPV of the mining operation over the specified time period. The objective function employs two continuous decision variables for each block $n$. The extraction variable $y^t_n$ represents the fraction of both ore and waste material from the block $n$ to be mined in period $t$, while the processing variable $x^t_n$ represents the fraction of ore from the block $n$ to be processed in period $t$. The use of continuous variables allows blocks to be fractionally extracted and processed across multiple time periods, providing flexibility in meeting capacity and quality constraints. 

\begin{equation} 
\text{Max}\sum^T_{t=1}\sum^N_{n=1}\left(\frac{v^t_n\times x^t_n - q^t_n\times y^t_n}{(1+r)^t}\right) 
\label{obj}
\end{equation}

 {This objective function is subject to the following constraints: \cref{equ: c1,equ: c2,equ: c3,equ: c4,equ: c5,equ: c6,equ: c7,equ: c8,equ: c9,equ: c10,equ: c11,equ: c12,equ: c13,equ: c14,equ: c15}

\subsubsection{Mining Capacity Constraints}
Mining capacity constraints ensure that the total volume of material extracted per period remains within operational equipment limits.~ \cref{equ: c1} enforces the standard upper extraction bound $Cu^t$. Constraints~\cref{equ: c2,equ: c3,equ: c4} introduce the novel dynamic lower bound mechanism designed specifically for end-of-mine-life scenarios. This mechanism relies on the remaining extractable fraction of each block, computed as: $re_n^t = (o_n + w_n) \times \left(1 - \sum_{j=1}^{t-1} y_n^j \right) \quad \forall n, \forall t$.
The auxiliary binary variable $l^t$ acts as a feasibility switch based on the total remaining reserves $\sum_{n=1}^N re_n^t$ available at the start of period $t$. It remains active ($l^t = 1$) only when total remaining reserves exceed the minimum threshold $Cl^t$. When reserves fall below this limit, the Big-M formulation in \cref{equ: c2,equ: c3} forces $l^t$ to zero, which in turn deactivates the lower bound in \cref{equ: c4}, allowing the operation to wind down naturally without triggering infeasibility. This prevents the solver from extending the mine life artificially or incurring unnecessary costs simply to satisfy a fixed minimum extraction floor that the remaining material cannot support.
\begin{equation}
\label{equ: c1}
\begin{gathered}
\sum^N_{n=1} (o_n + w_n) \times y_n^t \leq Cu^t
\quad \quad \quad \quad\forall_t \in {1,\dots, T};
\end{gathered}
\end{equation}
\begin{equation}
\label{equ: c2}
\begin{gathered}
l^t \le \left( 1 + \frac{\sum_{n=1}^N re_n^t - Cl^t}{M} \right)
\quad \quad \quad \quad \forall_t \in {1,\dots, T};
\end{gathered}
\end{equation}
\begin{equation}
\label{equ: c3}
\begin{gathered}
l^t \ge \left( \frac{\sum_{n=1}^N re_n^t - Cl^t}{M} \right)
\quad \quad \quad \quad \quad \quad \forall_t \in {1,\dots, T};
\end{gathered}
\end{equation}
\begin{equation}
\label{equ: c4}
\begin{gathered}
\sum^N_{n=1} (o_n + w_n) \times y_n^t \ge Cl^t \times l^t
\quad \quad \quad \quad \forall_t \in {1,\dots, T};
\end{gathered}
\end{equation}
\subsubsection{Processing capacity constraints}
Processing capacity constraints govern ore throughput through the mill, maintaining plant utilization within its designed operational range. Consistent mill utilization is critical for stable metallurgical recovery and minimizing disruptions to downstream operations. Because spatial grade variations may make simultaneous satisfaction of throughput and grade targets difficult, pre-stripping of waste is often required to ensure a uniform ore feed once production begins. \cref{equ: c5,equ: c6} define the upper and lower throughput bounds using the continuous processing variable $x_n^t$.
\begin{equation}
\label{equ: c5}
\begin{gathered}
\sum^N_{n=1}(o_n\times x^t_n)\leq Qu^t
\quad \quad \quad  \quad \quad \quad \quad \quad \forall_t \in {1,\dots,T};
\end{gathered}
\end{equation}
\begin{equation}
\label{equ: c6}
\begin{gathered}
\sum^N_{n=1}(o_n\times x^t_n)\geq Ql^t
\quad \quad \quad \quad \quad \quad \quad \quad \forall_t \in {1,\dots,T};
\end{gathered}
\end{equation}
\subsubsection{Grade Blending Constraints}
Grade blending constraints ensure that the ore feed delivered to the processing plant in each period satisfies the quality requirements necessary for efficient metallurgical recovery, consistent with long-term mine planning models that incorporate grade control~\cite{KOUSHAVAND2014451}. Processing plants are designed to operate optimally within a defined feed grade range, and significant deviations can result in reduced recovery rates or equipment damage. \cref{equ: c7,equ: c8} enforce these quality bounds by constraining the weighted average grade of the processed feed to lie within the acceptable interval [$\underline{g}^t, \bar{g}^t$] in each period, using the processing variable $x^t_n$ as the blending control mechanism.

\begin{equation}
\label{equ: c7}
\begin{gathered}
\sum^N_{n=1}(g_n - \bar{g}^t) \times (o_n\times x^t_n) \leq 0
\quad \quad \quad  \quad\forall_t \in {1,\dots,T};
\end{gathered}
\end{equation}
\begin{equation}
\label{equ: c8}
\begin{gathered}
\sum^N_{n=1}(g_n - \underline{g}^t) \times (o_n\times x^t_n) \geq 0 
\quad \quad \quad \quad \forall_t \in {1,\dots,T};
\end{gathered}
\end{equation}
\subsubsection{Block Precedence Constraints}
Block precedence constraints enforce slope stability by requiring that all overlying blocks are completely removed before any deeper block can be accessed. The primary contribution here is the explicit geometric construction of the predecessor set $H_n(D)$ from real slope angle requirements, rather than treating it as a theoretical given as many prior formulations do. Accessing a block at $(i, j, k)$ explicitly requires the prior removal of a defined cone of overlying blocks, including the block at $(i, j, k-1)$ and its necessary adjacent neighbors. The binary variable $b^t_n$ operationalizes this dependency, where \cref{equ: c9} prevents extraction of block $n$ from beginning in period $t$ unless every predecessor block $d \in H_n(D)$ has been fully extracted by that period. \cref{equ: c10} links the continuous extraction variable to the binary indicator to ensure consistency, and \cref{equ: c11} enforces temporal monotonicity so that once a block becomes accessible it remains accessible in all subsequent periods.
\begin{equation}
\label{equ: c9}
\begin{gathered}
b^t_n - \sum_{i=1}^{t} y^i_d \le 0, \quad \quad \quad  d \in H_n(D)  \quad\forall n \in {1,\ldots,N}, \quad \forall t \in {1,\ldots,T};
\end{gathered}
\end{equation}
\begin{equation}
\label{equ: c10}
\begin{gathered}
\sum_{i=1}^{t} y^i_n - b^t_n \le 0 
\quad \quad \quad \quad \quad \quad \quad \quad \quad \forall n \in {1,\ldots,N}, \quad \forall t \in {1,\ldots,T};
\end{gathered}
\end{equation}
\begin{equation}
\label{equ: c11}
\begin{gathered}
b^t_n - b^{t+1}_n \le 0 
\quad \quad \quad \quad \quad \quad \quad \quad\forall n \in {1,\ldots,N}, \quad \forall t \in {1,\ldots,T-1} ;
\end{gathered}
\end{equation}
\subsubsection{Variable control constraints}
Variable control ensures material conservation and enforces strict extraction processing coupling that reflects real world operations.~\cref{equ: c12,equ: c13} are the  constraints which guarantee that the total fraction of each block mined or processed across all scheduling periods does not exceed one. A key contribution is the auxiliary binary $u_n^t$ in~\cref{equ: c14,equ: c15}, which acts as an intermediate linkage variable to ensure processing only begins once a block has been fully extracted. This resolves limitations in previous formulations where partial extraction permitted immediate processing, a physical impossibility since a block must be fully blasted and excavated before its ore can be separated.
\begin{equation}
\label{equ: c12}
\begin{gathered}
\sum^T_{t=1}y_n^t\leq 1 
\quad \quad \quad \quad \quad 
\quad \quad \quad \quad \quad \quad \forall_n \in{1,\dots,N};
\end{gathered}
\end{equation}
\begin{equation}
\label{equ: c13}
\begin{gathered}
\sum^T_{t=1}x_n^t\leq 1 
\quad \quad \quad \quad \quad 
\quad \quad \quad \quad \quad \quad \forall_n \in{1,\dots,N};
\end{gathered}
\end{equation}
\begin{equation}
\label{equ: c14}
\begin{gathered}
u_n^t \le \sum_{j=1}^t y_n^j
\quad \quad \quad \quad \forall_n \in{1,\dots,N}; \quad \forall_t \in{1,\dots,T};
\end{gathered}
\end{equation}
\begin{equation}
\label{equ: c15}
\begin{gathered}
x_n^t \le u_n^t 
\quad \quad \quad \quad \quad\forall_n \in{1,\dots,N}; \quad \forall_t \in{1,\dots,T};
\end{gathered}
\end{equation}

\section{Performance Analysis}
\label{sec:result}
This section presents experimental results obtained from a diverse suite of mining instances that vary in block counts and time periods. A comparative analysis of NPV and optimality gaps is provided, along with Gantt chart visualizations of the extraction sequence and an assessment of model robustness. Numerical performance and computational efficiency are evaluated with particular attention to execution time and the impact of the LLM prompt refresh rate on scalability.

All methods are evaluated both without domain inputs and with the provision of context-aware mining data.

\subsection{Experimental Setup}
The experimental framework is established following standard practices in open-pit mining optimization. All blocks are configured with a uniform mass of 50 tons to ensure consistency across evaluations. This standardized block size is used for both the MILP model evaluation and the simulator-LLM interaction framework.
The economic model uses an ore processing revenue of \$100 per ton and a processing cost of \$10 per ton, reflecting typical market conditions in 
mining operations. A discount rate of $r = 0.10$ (10\%)~\cite{fontes2021analysis} is applied to account for the time value of money across planning periods.
The mining and processing capacities are configured to reflect operational constraints. The mining capacity ranges from 20 to 70 tons per period, while the processing capacity ranges from 10 to 40 tons per period. 
\subsubsection{MILP Baseline Configuration}
The MILP formulation was solved using the Gurobi Optimizer (version 13.0.0)~\cite{gurobi} via the Python interface. The solver was executed on a computing environment equipped with $16$ logical CPU cores and $94$ GB of RAM. For small instances ($\le 20$ blocks), default solver tolerances were applied to ensure high-precision optimality. For
larger instances ($> 20$ blocks), where combinatorial complexity increases substantially, the relative optimality tolerance was set to 0.05 (5\%)~\cite{loor2020applying} to ensure timely convergence.

\subsubsection{LLM Experimental Setup}
\label{subsec:llm_setup}
Two open-source LLMs are evaluated to assess the generalizability of the simulator-based scheduling framework across models of varying architectures and parameter scales. GPT-OSS~\cite{gptoss} (20B parameters), and DeepSeek-R1~\cite{deepseekr1} (14B parameters) are utilized via the Ollama framework~\cite{ollama}, enabling unrestricted local execution without reliance on external API services. This local deployment ensures that all mine data remains within the operational environment throughout experimentation, consistent with the closed-system design of the proposed framework. Rather than relying on unstructured text generation, a structured prompting framework is engineered for all models, consisting of a static system instruction and dynamic iterative updates.

\begin{figure}
\centering
\begin{tcolorbox}[colback=gray!5, colframe=black, title=\textbf{LLM System Instruction}]
\small
You are an expert mine scheduler. Your sole objective is to maximize the Net Present Value (NPV) of a mining operation by selecting the optimal action at each step. At every step, I will provide:
\ul{Mine State: A 3D array representing layers. Each element is a tuple: \texttt{(block\_number, ore\_concentration, tons\_to\_extract, tons\_to\_process)}.}
Possible Actions: A list of available actions. You must choose exactly one.

Action Types:

Extraction (Choice 1): \texttt{(1, block\_number, max\_extractable, npv\_impact)}. Incurs cost. Amount must be $> 0$ and $\leq$ max.

Processing (Choice 2): \texttt{(2, block\_number, min\_blend, max\_blend, npv\_impact)}. Generates revenue. Amount must be between \texttt{min\_blend} and \texttt{max\_blend}.

Constraints \& Logic:

Dependencies: Processing is only available after full extraction.

Time: Time periods advance automatically based on capacity limits.

Output Format: Respond ONLY with a valid JSON object:

\texttt{\{ "choice": [1 or 2], "block\_number": (int), "amount": (int) \}}

\end{tcolorbox}
\caption{System instruction defining the LLM scheduling rules, action space, and output format.}
\label{fig:llm_prompt_refined}
\Description{Screenshot of a structured text prompt or system instruction box. The content defines three sections: the LLM agent's scheduling rules, the available action space describing the set of permissible scheduling decisions, and the required output format specifying how the agent should return its schedule.}
\end{figure}

As shown in~\cref{fig:llm_prompt_refined}, the system instruction casts the LLM as an expert mine scheduler tasked solely with maximizing NPV. It defines two action types: extraction (Action 1), which incurs cost with an amount bounded above by the block maximum, and processing (Action 2), which generates revenue with an amount bounded by blending limits. The instruction enforces the core dependency rule that processing becomes available only after full extraction, mandates extraction whenever feasible to prevent the model stalling on early negative cash flows, and requires output strictly as a JSON object to ensure seamless simulator parsing. The same system instruction is applied uniformly across all models to ensure that observed differences in scheduling performance are attributable to the reasoning capabilities of each model rather than differences in prompt design.

During execution, the simulator drives the LLM via dynamic iterative prompts. As illustrated in~\cref{fig:llm_prompts}, at $t=1$ the model receives the full mine state and a filtered list of feasible actions with their immediate NPV impacts. From $t=2$ onward, the prompt is prepended with feedback on the prior action and the updated cumulative NPV, creating a closed feedback loop that adapts LLM decisions to the evolving mine state without requiring full re-optimization.
\begin{figure}
    \centering
    \begin{tcolorbox}[colback=gray!5, colframe=black, title=\textbf{LLM Prompt for each iteration}]
    \small
    \textbf{Simulator (Initial Iteration $t=1$):}
Time\_period-1: \ul{Here is Mine State (Truncated): [[[(1, 0.66, 50, 33), (2, 0.67, 50, 33), ... [(10, 0.63, 50, 31), (11, 0.74, 50, 37), (12, 0.75, 50, 37)], ... (15, 0.66, 50, 33)]],[[(16, 0.58, 50, 28), (17, 0.66, 50, 33), ... (30, 0.67, 50, 33)]]]}

Here are the possible actions (Truncated): [(1, 1, 50, -9.09), (1, 2, 50, -9.09), ... (1, 12, 50, -9.09), ... (1, 15, 50, -9.09)]
\smallskip
\par\noindent\rule{\textwidth}{0.4pt}
\smallskip
\textbf{Simulator (Iteration $t=2,3,4...$):}

choice was: 1, block\_number was: 12, amount was: 50. Current NPV is: -454.55

Time\_period-1: \ul{Here is Mine State (Truncated): [[[(1, 0.66, 50, 33), (2, 0.67, 50, 33), ... [(10, 0.63, 50, 31), (11, 0.74, 50, 37), (12, 0.75, 0, 37)], ... (15, 0.66, 50, 33)]],[[(16, 0.58, 50, 28), (17, 0.66, 50, 33), ... (30, 0.67, 50, 33)]]]}

Here are the possible actions (Truncated): [(1, 1, 50, -9.09), ... (1, 15, 50, -9.09), (2, 12, 0.0, 37, 90.91)]
\end{tcolorbox}
\caption{Iterative prompts showing the initial mine state at $t=1$ and the updated state with prior action feedback at $t=2,3,4$.}
\label{fig:llm_prompts}
\end{figure}

\subsection{Results}
The comparative performance of the deterministic MILP model and the LLM-based simulator is evaluated both with and without contextual information, alongside standard Random and Greedy heuristic baselines. 
Detailed visualizations are presented to illustrate the impact of parameter variations on the overall performance of the MILP formulation.

\subsubsection{NPV Gain and Optimality Gap}
This section provides a visual assessment and comparative analysis of the LLM-simulator approach against the MILP baseline and heuristic benchmarks.

\begin{figure}
    \centering
    \includegraphics[width=.6\linewidth]{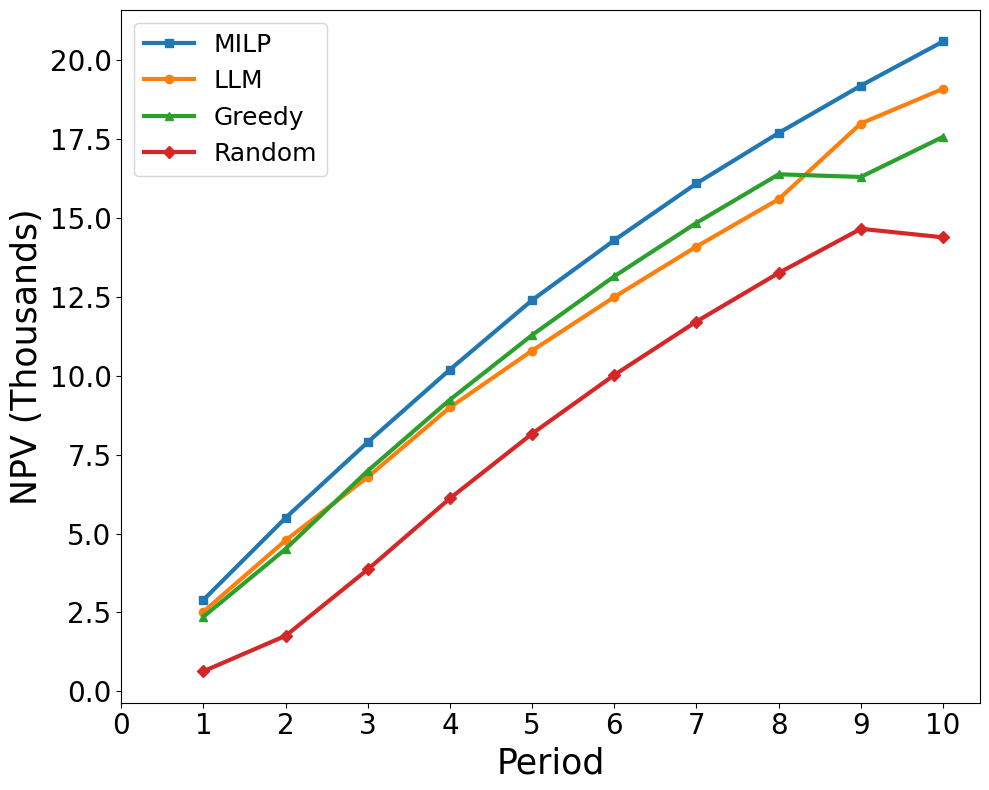}
    \caption{Comparative NPV for a 27-block mining instance across MILP, the
    LLM-based simulator, greedy, and random strategies.}
    \label{fig:ogap}
    \Description{Bar chart comparing NPV outcomes across four scheduling
    strategies for a 27-block mining instance. MILP serves as the optimal
    reference.}
\end{figure}

\cref{fig:ogap} compares all scheduling strategies across a 27-block mine 
instance over 10 time periods, with MILP serving as the optimal reference. 
To quantify deviation from this theoretical maximum, the optimality gap is 
defined as
\begin{equation}
    \mathrm{OGap}(t) = \frac{NPV_{M}(t) - NPV_{s}(t)}{NPV_{M}(t)},
\end{equation}
where $NPV_{M}(t)$ denotes the NPV obtained by the MILP solution and $NPV_{s}(t)$ denotes the NPV produced by the evaluated strategy at period $t$. Among the LLM-based approaches, GPT-OSS was selected as the representative model for this comparison, as it consistently retained higher NPV values across planning periods relative to DeepSeek, demonstrating stronger alignment with the MILP benchmark under equivalent contextual inputs.

\begin{figure}
    \centering
    \includegraphics[width=.6\linewidth]{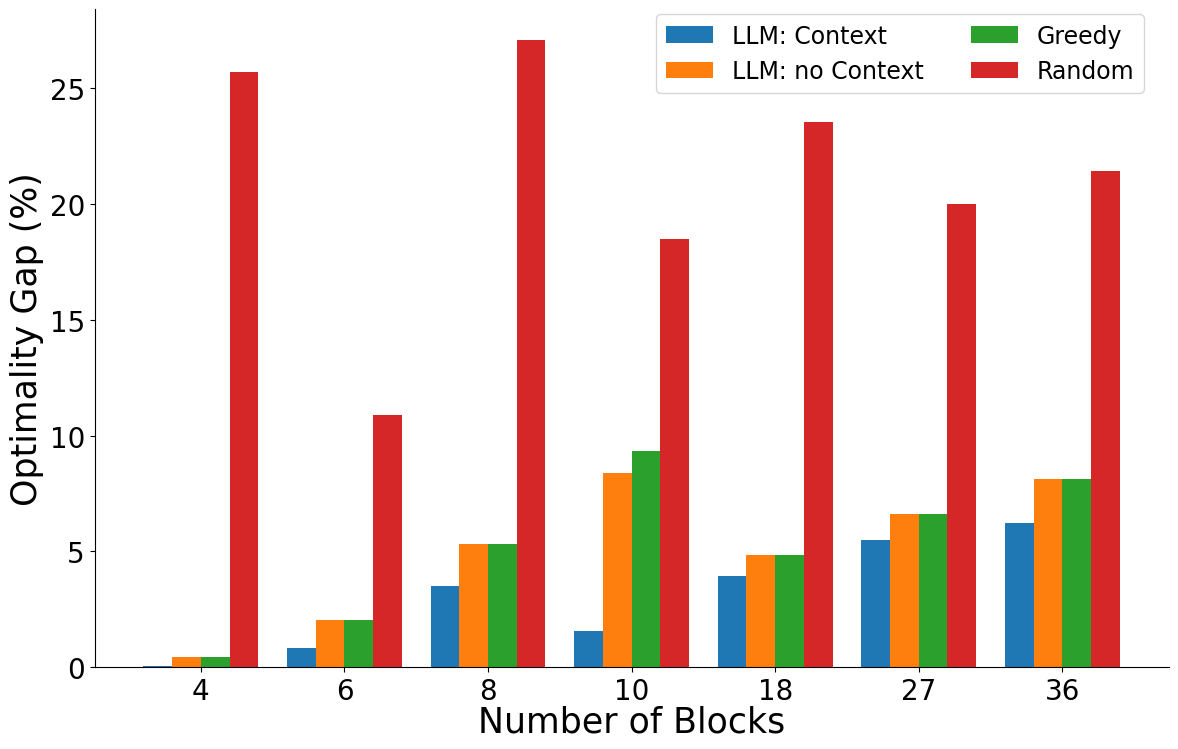}
    \caption{Optimality gap (lower is better) for each strategy relative to
    MILP across varying numbers of blocks.}
    \label{fig: comparison}
    \Description{Grouped bar chart comparing NPV optimality gaps for LLM,
    greedy, and random strategies against the MILP baseline over varying
    problem sizes.}
\end{figure}
\cref{fig: comparison} illustrates the optimality gap of the solution strategies in comparison to the MILP benchmark over time periods. The results for the first four groups correspond to single-layer mine configurations without precedence constraints, while the results for later groups correspond to multi-layer settings where precedence constraints are applied.

The random strategy yields the largest optimality gap, confirming that uncoordinated scheduling produces far-from-optimal results. The greedy and no-context LLM approaches achieve similar intermediate NPV values. The context-aware LLM, provided with live mine-state data, narrows the gap substantially and approaches the MILP optimum.

\subsubsection{Numerical Results and Comparative Analysis}
This subsection summarizes the experimental results and provides a comparative numerical analysis of the performance of each approach. \cref{tab:method_comparison} compares scheduling metrics across three planning periods for five strategies. MILP achieves the highest NPV of \$20{,}578.86 by period~10 and the greatest average extraction tonnage (42.31 tons), reflecting a highly selective strategy that prioritizes the most resource-rich blocks.
\begin{table*}
\caption{Comparison of Operation Metrics over Different Periods}
\label{tab:method_comparison}
\centering
\small
\begin{tabular*}{\textwidth}{@{\extracolsep{\fill}}l c  c c c c r}
\toprule
\multirow{2}{*}
{\textbf{Method}} & 
\multirow{2}{*}
{\textbf{Period}} & \multicolumn{2}{c}{\textbf{No. of Operations}} & \multicolumn{2}{c}{\textbf{Avg. Tonnage}} & 
\multirow{2}{*}
{\textbf{NPV}} \\
\cmidrule(lr){3-4} \cmidrule(lr){5-6}
 & & \textbf{Ext.} & \textbf{Proc.} & \textbf{Ext.} & \textbf{Proc.} & \\
\midrule

\multirow{3}{*}{MILP}
 & 1 & 1 & 1 & 50.00 & 37.00 & 2909.00 \\
 & 5 & 8 & 8 & 37.50 & 24.25 & 12340.63 \\
 & 10 & 13 & 18 & 42.31 & 21.89 & 20578.86 \\
\midrule

\multirow{3}{*}{\shortstack[l]{LLM \\ (Context)}}
 & 1 & 2 & 1 & 35.00 & 37.00 & 2727.28 \\
 & 5 & 14 & 7 & 25.00 & 27.14 &  11688.44 \\
 & 10 & 27 & 16 & 25.93 & 24.38 & 19455.93 \\
\midrule

\multirow{3}{*}{\shortstack[l]{LLM \\ (No Context)}}
 & 1 & 2 & 1 & 35.00 & 33.00 & 2363.63 \\
 & 5 & 11 & 8 & 31.82 & 23.25 & 11294.74 \\
 & 10 & 22 & 19 & 31.82 & 20.32 & 19062.21 \\
\midrule

\multirow{3}{*}{Greedy}
 & 1 & 2 & 1 & 35.00 & 33.00 & 2363.63 \\
 & 5 & 11 & 8 & 31.82 & 23.25 & 11294.73 \\
 & 10 & 22 & 19 & 31.82 & 20.32 & 19062.21 \\
\midrule

\multirow{3}{*}{Random} 
 & 1 & 9 & 0 & 7.78 & 0.00 & -636.36 \\ 
 & 5 & 16 & 9 & 21.88 & 13.33 & 5620.03 \\ 
 & 10 & 30 & 34 & 23.33 & 9.41 & 13387.50 \\
\bottomrule
\end{tabular*}
\end{table*}
The context-aware LLM proves to be a robust surrogate for mathematical optimization. By Period~10 it achieves an NPV of \$19{,}455.93, capturing approximately 94.5\% of the MILP optimum. Although it executes more operations than MILP (43 versus 31), access to live mine-state data allows it to meaningfully outperform both the no-context LLM and the Greedy baseline, which plateau at \$19{,}062.21.

The Greedy and no-context LLM strategies exhibit identical performance trajectories, suggesting that without environmental context an LLM reverts to simple short-term profit-seeking heuristics indistinguishable from greedy search. The Random strategy highlights the cost of uncoordinated activity, as despite the highest operation count (64 total), it yields the lowest NPV (\$13{,}387.50) due to poor prioritization and low average tonnage per operation.

\subsubsection{Efficiency and Scalability}

This subsection characterizes the temporal cost of achieving higher optimality. Due to the vast execution speed differences between heuristic and optimization models, a logarithmic scale is employed to visualize the orders of magnitude in computational effort.
\begin{figure}
\centering
\subfloat[Block size = 18]{\includegraphics[width=.5\linewidth]{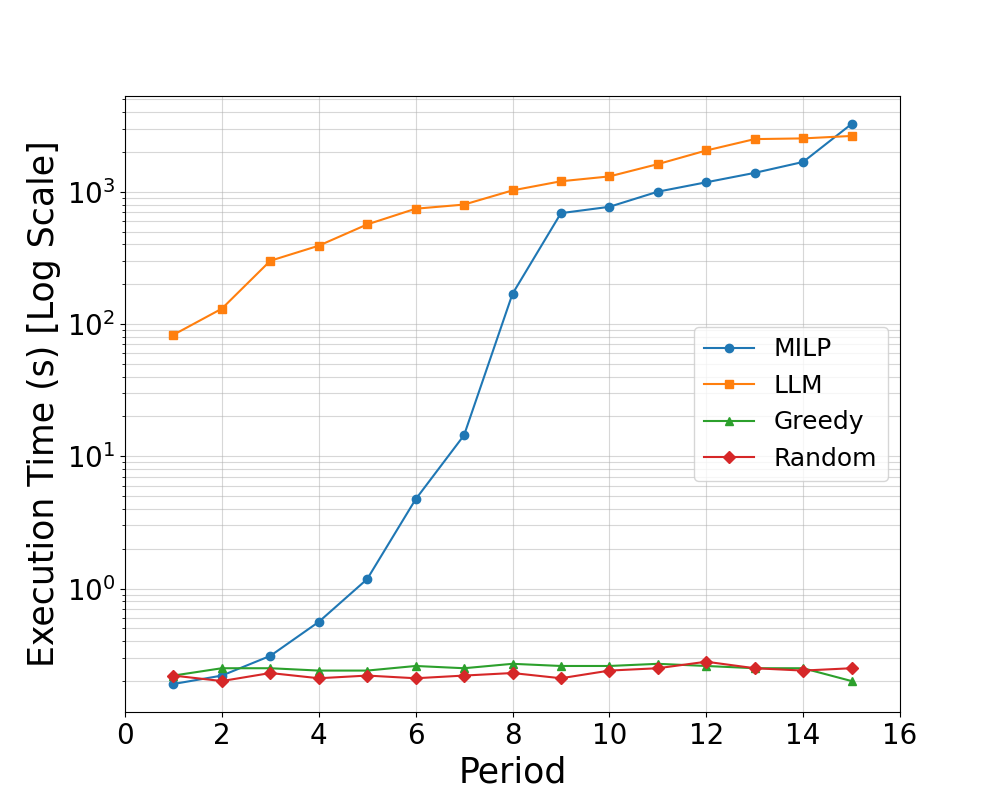}\label{fig:pvst_a}}\hfill
\subfloat[Block size = 45]{\includegraphics[width=.5\linewidth]{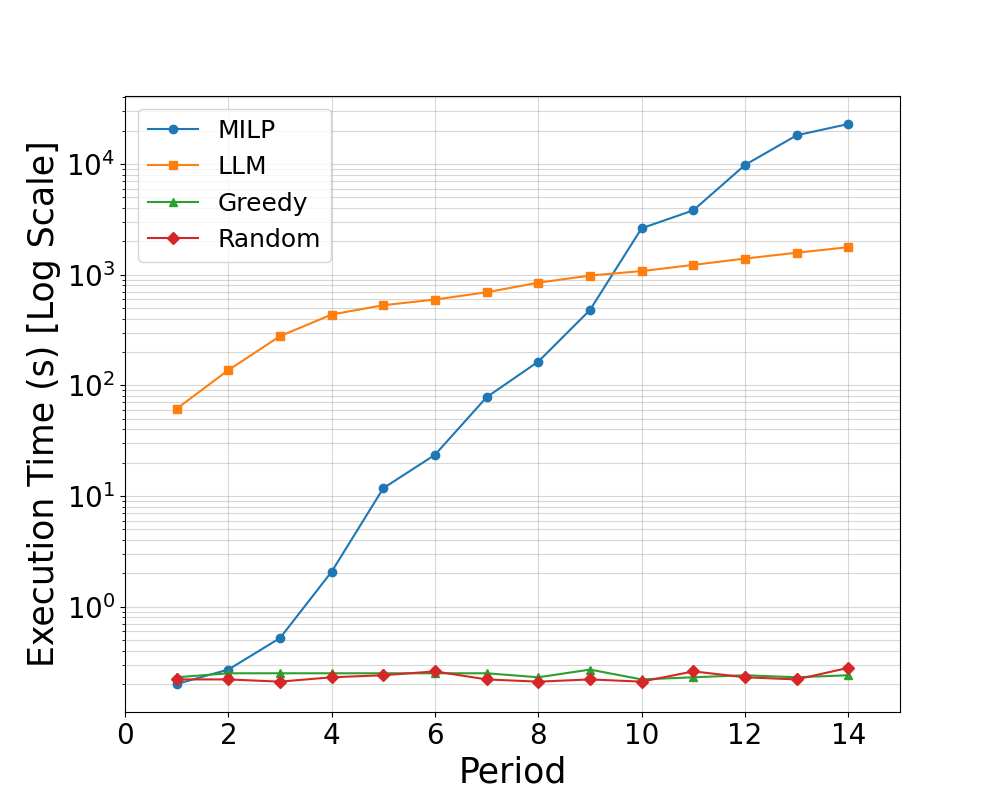}\label{fig:pvst_b}}
\caption{Execution times to reach convergence (seconds, log scale).}
\label{fig:pvst}
\Description{Line chart displaying execution times on a logarithmic scale, allowing comparison of computational cost across methods as the time period grows.}
\end{figure}

As illustrated in ~\cref{fig:pvst}, random and greedy heuristics require the least effort due to low algorithmic complexity. MILP is efficient on small
instances but exhibits a sharp exponential increase in execution time as the
time period expands, reflecting the vast combinatorial search space.
The LLM-based approach initially shows higher latency but grows at a more
linear and stable rate, indicating superior scalability for large-scale problems.

\subsubsection{Computational Trade-Off}

\cref{tab:method_compare} examines the trade-off between computational overhead and solution quality across instance sizes $N$ over $T=15$ periods. Both LLM-based approaches scale linearly and remain well within the MILP solution quality at every tested instance size. GPT-OSS achieves 99.75\% of the reported MILP NPV at $N=36$ and stays within 1.2\% of the MILP maximum at $N=45$ while requiring only 15\% of the computation time. DeepSeek matches GPT-OSS solution quality at $N=18$ and $N=27$ but yields a lower NPV at $N=45$, indicating reduced robustness at larger scales. Despite having fewer parameters, DeepSeek incurs consistently higher execution times across all instance sizes, attributable to its chain-of-thought reasoning architecture, which generates lengthy intermediate reasoning tokens before producing a final action selection.
\begin{table*}
    \centering
    \caption{NPV and execution time across varying instance sizes ($N$).}
    \label{tab:method_compare}
    \resizebox{\textwidth}{!}{
    \begin{tabular}{l l cc cc cc cc}
        \toprule
        \multirow{2}{*}{\textbf{Method}} & \multirow{2}{*}{\textbf{Complexity}} & \multicolumn{2}{c}{\textbf{N=18}} & \multicolumn{2}{c}{\textbf{N=27}} & \multicolumn{2}{c}{\textbf{N=36}} & \multicolumn{2}{c}{\textbf{N=45}} \\
        \cmidrule(lr){3-4} \cmidrule(lr){5-6} \cmidrule(lr){7-8} \cmidrule(lr){9-10}
        & & \textbf{NPV} & \textbf{Time (s)} & \textbf{NPV} & \textbf{Time (s)} & \textbf{NPV} & \textbf{Time (s)} & \textbf{NPV} & \textbf{Time (s)} \\
        \midrule
        MILP              & $O(2^N)$ & 24650.41 & 3247.72 & 25694.16\textsuperscript{\dag} & 437.31 & 25694.16\textsuperscript{\dag} & 3200.26 & 25664.32\textsuperscript{\dag} & 12603.30 \\
        GPT-OSS (Context)     & $O(N)$   & 24083.20 & 1527.55 & 24278.91 & 1313.86 & 25630.80 & 1619.67 & 25369.34 & 1880.35 \\
        DeepSeek (Context)     & $O(N)$   & 24083.20 & 2827.75 & 24278.91 & 3157.14 & 25630.80 & 3249.24 & 24687.95 & 3541.30 \\
        \bottomrule
    \end{tabular}
    }
    \begin{flushleft}
    \footnotesize{\textsuperscript{\dag} Solver terminated with a 5-6\% optimality gap. Time(s) represents the total computation time.}
    \end{flushleft}
\end{table*}
While MILP provides the theoretical global optimum, its application is constrained by exponential complexity $O(2^N)$. For $N=45$ the solver requires over 12{,}000~seconds despite accepting a 5--6\% termination gap. The lower time at $N=27$ results from the rapid identification of an incumbent solution satisfying the tolerance threshold. These results confirm that LLM-based methods can effectively approximate MILP strategic decision-making at a fraction of the runtime, though model capacity meaningfully influences both solution quality and computational efficiency.

\subsubsection{Extraction Sequence Visualization}
To validate the solution quality of the LLM-based approaches, Gantt charts for MILP, GPT-OSS, and DeepSeek are presented in~\cref{fig:gantt_all} for the 27-block instance. The MILP schedule is sparser, reflecting the solver's value-driven sequencing logic. Notably, GPT-OSS and DeepSeek produce identical schedules for this instance, with both models arriving at the same block ordering and phase transitions across all 15 periods despite being prompted independently. This convergence between two architecturally distinct models suggests that the LLM-based framework consistently recovers a coherent mining strategy. The overall structure of both LLM schedules closely mirrors the MILP benchmark in sequencing priorities, providing qualitative evidence that the models replicate the underlying operational logic without exhaustive combinatorial search.
\begin{figure}
    \centering
    \subfloat[MILP Benchmark Schedule]{%
        \includegraphics[width=0.33\linewidth,keepaspectratio]{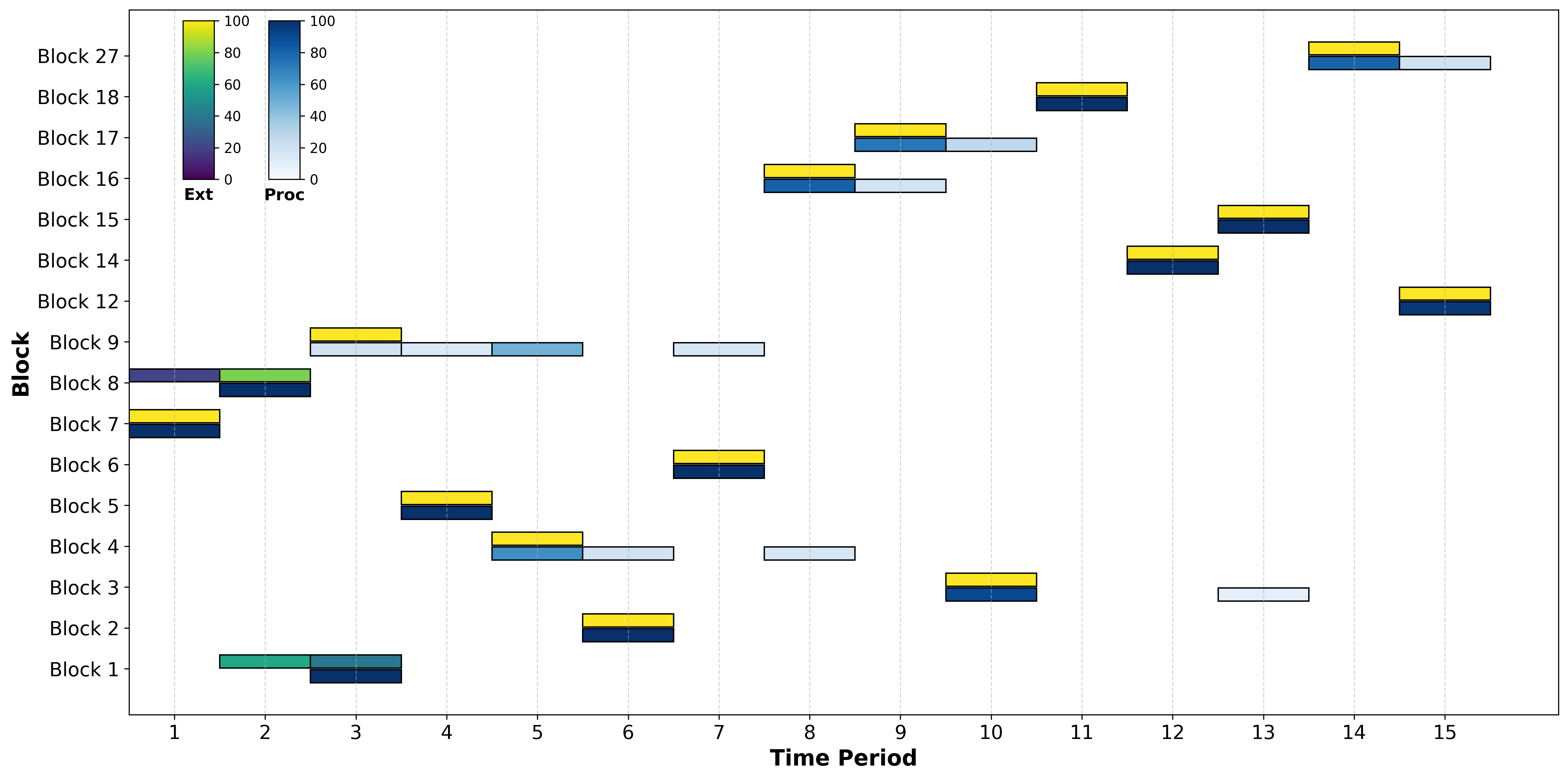}%
        \label{fig:gantt_milp}}\hfill
    \subfloat[GPT-OSS Schedule]{%
        \includegraphics[width=0.33\linewidth,keepaspectratio]{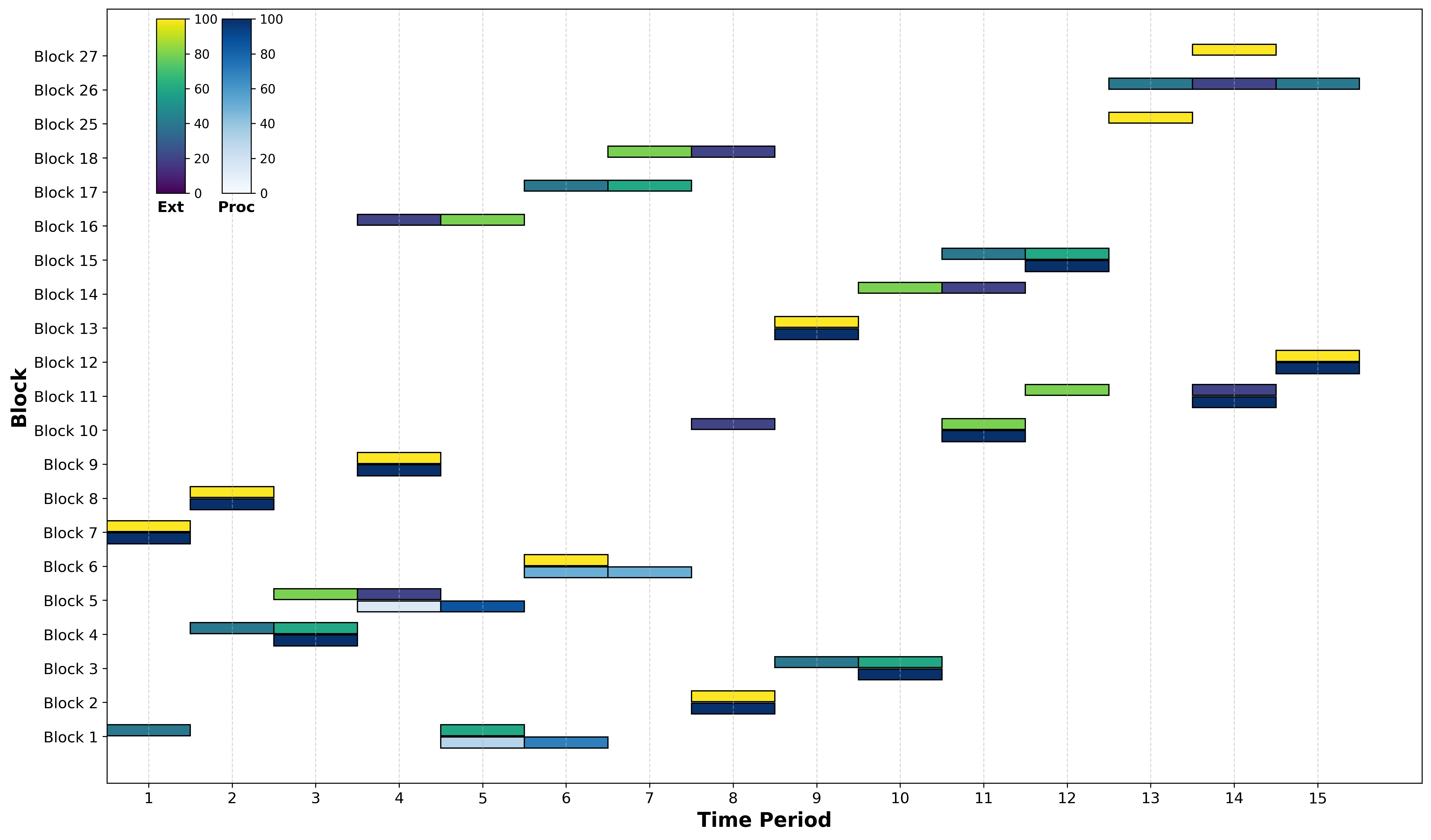}%
        \label{fig:gantt_gpt}}\hfill
    \subfloat[DeepSeek Schedule]{%
        \includegraphics[width=0.33\linewidth,keepaspectratio]{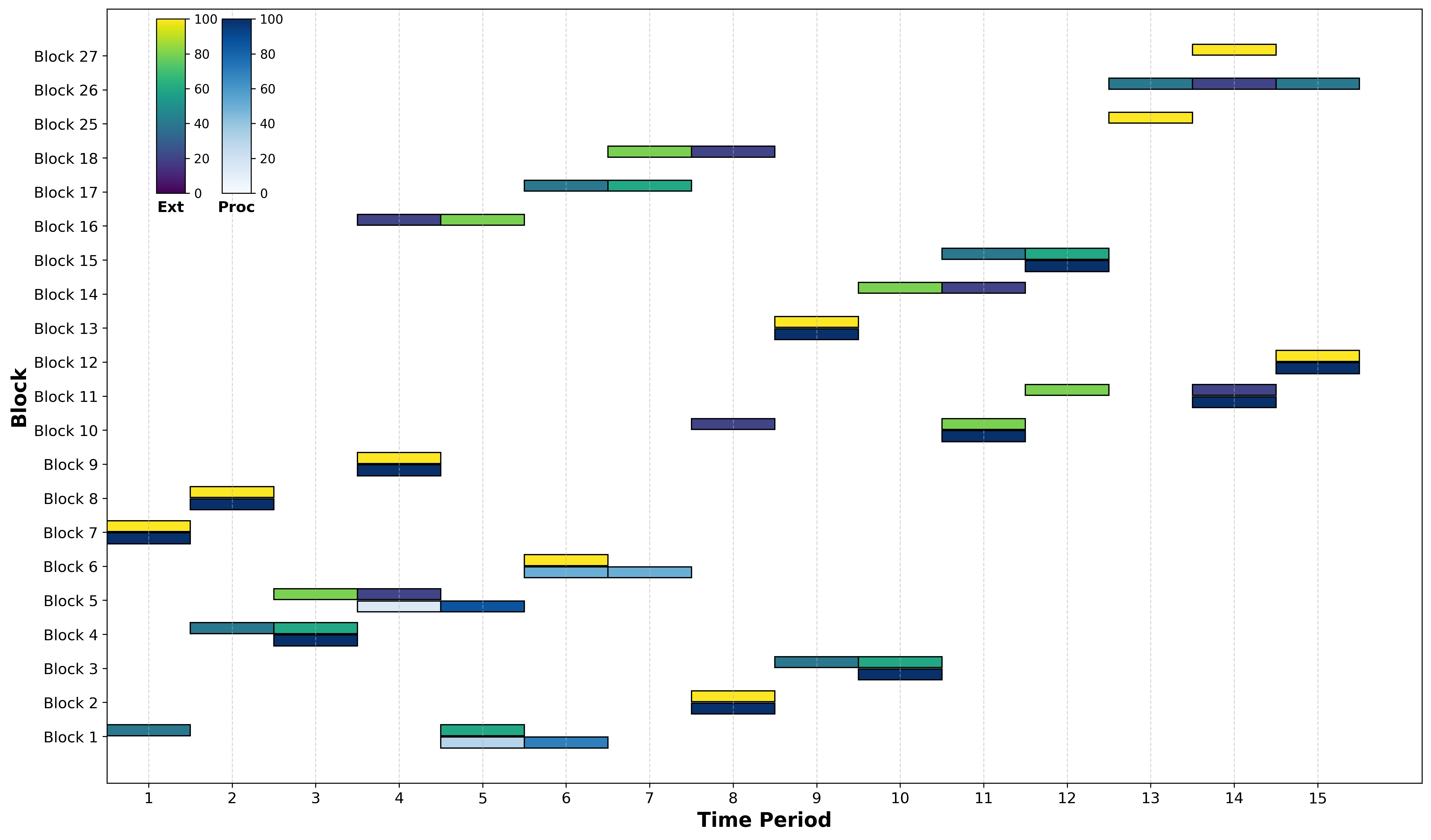}%
        \label{fig:gantt_deepseek}}
    \caption{Gantt chart comparison of mining schedules for the 27-block
    instance across MILP, GPT-OSS, and DeepSeek. Each bar represents a
    block's extraction (Ext) and processing (Proc) phases, coloured by
    completion percentage.}
    \label{fig:gantt_all}
    \Description{Three Gantt charts comparing block extraction and processing
    schedules over 15 time periods for a 27-block mining instance.}
\end{figure}
\subsubsection{LLM Agent Transparency and Operational Output}
Operational transparency is a key advantage of the simulator-LLM framework over conventional exact solvers. Unlike batch optimization solvers that yield no usable output until solving completes, the simulator emits a structured dispatch log at each planning step, translating raw LLM decisions directly into interpretable execution directives that mine planners can monitor and validate in real time.

\begin{figure}
\centering
\begin{lstlisting}[basicstyle=\ttfamily\scriptsize, frame=single, breaklines=true, columns=fullflexible, keepspaces=true]
====================================
SIMULATOR DISPATCH LOG
====================================
[RAW LLM OUTPUT]  
Time-period 1: (1, 12, 50)  
Time-period 1: (2, 12, 37)  
Time-period 1: (1, 11, 20)
Time-period 2: (1, 11, 30)
[PARSED EXECUTION DIRECTIVE]
> Initiating Period 1 Operations...
-- BLOCK 12 --
   Action : EXTRACTION
   Amount : 50.00 units
   Action : PROCESSING
   Amount : 37.00 units
-- BLOCK 11 --
   Action : EXTRACTION
   Amount : 20.00 units
Period 1 Dispatch Complete. No further actions viable.
Advancing to Period 2..
-- BLOCK 11 --
   Action : EXTRACTION
   Amount : 30.00 units
====================================
\end{lstlisting}

\caption{Console-style visualization demonstrating how the raw sequence arrays generated by the LLM are parsed into interpretable, real-time operational directives.}
\label{fig:console_output}
\end{figure}

\cref{fig:console_output} illustrates this process during execution. Each LLM output tuple encodes an action type (1~for extraction, 2~for processing), block ID, and material amount. For Block~12, the agent schedules 50~units for extraction followed immediately by 37~units for processing, demonstrating strict extraction-processing coupling. A fractional extraction of 20~units is then scheduled for Block~11. Because this exhausts viable capacity for Period~1, the simulator automatically advances to Period~2, where the agent schedules the remaining 30~units. This step-by-step visibility allows mine planners to verify physical feasibility, capacity constraints, and economic logic at every decision point.

\subsubsection{Effect of Context Refresh Rate}
This subsection investigates the sensitivity of the LLM optimizer to the context window refresh rate. Because the local LLM implementation lacks native long-term memory, context is managed by feeding sequential simulation results back into the prompt. A refresh size of~1 provides only the most recent state (high-frequency refresh), while ``Full'' retains the entire interaction history.

As illustrated in~\cref{fig:llmp}, a refresh size of~1 yields the highest NPV of \$12{,}455.24. Retaining excessive historical context appears to introduce noise and dilutes focus on the immediate optimization constraints. As the retained history grows toward ``Full'', NPV declines, indicating that the most recent mine state is the primary driver of decision-making accuracy in this setting. This behaviour also suggests that the LLM-based approach can scale efficiently to large mines: by focusing on local state variables rather than a growing global history, it avoids the exponential computational decay characteristic of traditional solvers.

\begin{figure}
\centering
\includegraphics[width=0.7\linewidth]{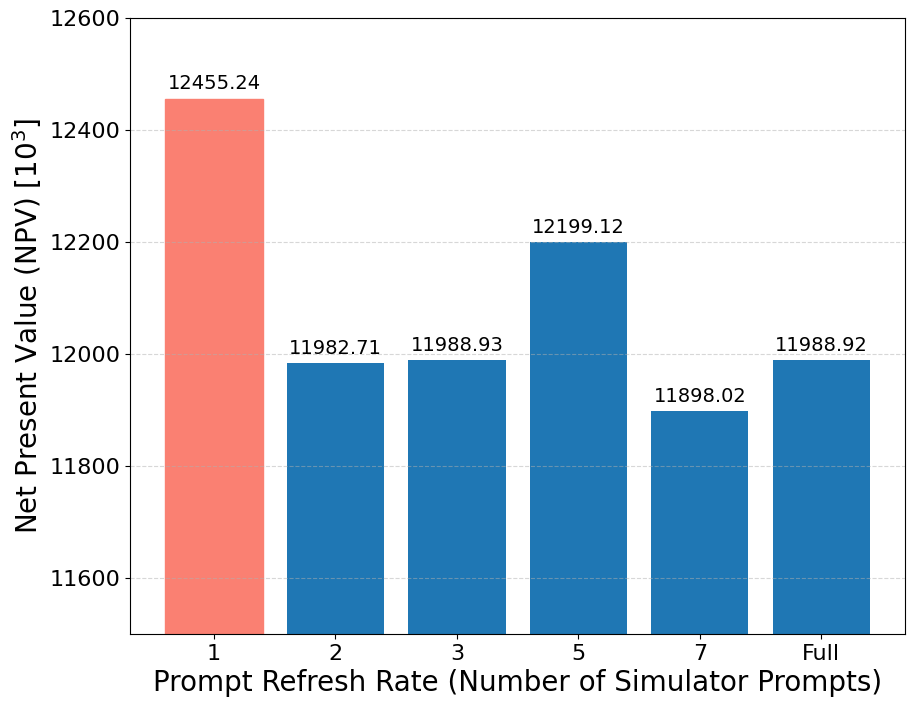}
\caption{Impact of Context Window Refresh Rate on LLM Solution Quality}
\label{fig:llmp}
\Description{Bar chart showing NPV values achieved by the GPT-OSS model under 
varying context window refresh rates, ranging from a prompt size of 1 (most 
recent state only) to Full (entire interaction history retained). The chart 
illustrates that a refresh size of 1 yields the highest NPV, with solution 
quality declining or stabilizing as more historical context is retained, 
suggesting that recent state information is the primary driver of scheduling 
accuracy.}
\end{figure}

\subsubsection{Method Selection Under Operational Requirements}
The choice of scheduling method depends on the operational demands of the mining environment, as summarized in~\cref{tab:decision_matrix}. MILP follows an all-or-nothing approach, evaluating all decision variables and constraints before producing a valid plan. This guarantees a global optimum but produces a planning blackout period during which no interim decisions can be made, and computational time grows exponentially as the search space expands.

\begin{table}
\centering
\caption{Strategic Comparison of Mining Methods}
\label{tab:decision_matrix}
\begin{tabular}{@{}llll@{}}
\toprule
\textbf{Feature} & \textbf{Greedy/Random} & \textbf{MILP} & \textbf{LLM-based} \\ \midrule
\textbf{\makecell[l]{Decision\\Process}} & Instantaneous & Batch-processed & Iterative/Adaptive \\
\textbf{\makecell[l]{Runtime\\Intervention}} & Yes & No & \makecell[l]{Yes\\(Real-time)} \\
\textbf{Optimality} & Non-optimal & \makecell[l]{Global\\Optimum} & Near-Optimal \\
\textbf{Scalability} & High & \makecell[l]{Low\\(Exponential)} & \makecell[l]{High\\(Linear)} \\
\textbf{\makecell[l]{Operational\\Flexibility}} & None & \makecell[l]{Static\\Constraints} & \makecell[l]{Dynamic\\Reasoning} \\
\bottomrule
\end{tabular}
\end{table}

The LLM-based framework operates within a continuous conversational loop with the simulator, issuing actionable extraction decisions at every step. This enables real-time monitoring and mid-process adjustments that are unattainable with MILP's static batch processing. As mine complexity grows, MILP eventually hits a computational wall, whereas the LLM sustains a consistent execution rate. For large-scale, dynamic operations where real-time visibility and scalability take precedence over absolute mathematical optimality, the LLM-based architecture presented here offers a compelling and practical alternative.
\section{Conclusion}
\label{sec:conclusion}
This work presented a simulator-driven LLM framework for autonomous open-pit production scheduling, evaluated against a rigorous MILP baseline and two heuristic benchmarks across a range of instance sizes and time periods. The central contribution is a closed, data-secure scheduling architecture in which an LLM operating zero-shot within a custom iterative simulator can recover near-optimal schedules without any domain-specific training, fine-tuning, or access to a global mathematical formulation. By delegating all geotechnical precedence checks, extraction-processing coupling, and capacity feasibility to the simulator, the LLM is relieved of constraint verification entirely and can direct its reasoning capacity toward long-term value maximization. This separation is the primary reason the framework achieves 94–99\% of MILP-optimal NPV while sustaining linear computational growth across all tested instance sizes.
To enable a trustworthy performance benchmark, a novel MILP formulation was 
developed that resolves three critical oversimplifications present in 
traditional models. A dynamic lower bound mechanism prevents mathematical 
infeasibility near end-of-mine-life by scaling minimum extraction thresholds 
to actual remaining reserves. Geometric construction of the full 3D 
predecessor set from slope angle requirements encodes realistic geotechnical 
safety directly into the constraint structure. Auxiliary binary variables 
enforce strict extraction-processing coupling, ensuring that no ore is routed 
to the processing plant before its host block has been completely excavated. 
Together, these contributions produce a benchmark that mine planners can 
genuinely trust. 
From an operational perspective, the framework's most distinctive advantage over both MILP and classical heuristics is its real-time interpretability. At every decision step, the simulator emits a structured dispatch log translating raw LLM outputs into immediately executable mine directives, enabling planners to monitor, audit, and intervene in the scheduling process without waiting for a batch solver to complete. This continuous operational visibility is essential for large-scale mining environments where equipment states, ore grades, and market conditions can shift mid-schedule.
The experimental results further demonstrate that context enrichment is decisive. When provided with live mine-state data, the LLM meaningfully outperforms both the no-context LLM and the Greedy baseline, confirming that real-time environmental feedback is the primary driver of LLM decision quality in this setting. The context refresh rate analysis shows that focusing on the most recent mine state rather than accumulating full interaction history maximizes solution quality while keeping memory requirements bounded, supporting deployment on resource-constrained industrial hardware and suggesting the framework can scale to very large instances without the computational decay characteristic of memory-intensive optimization approaches.
Future work will extend evaluation to larger real-world block models, incorporate geological grade uncertainty and commodity price variability into the simulator state, and investigate structured prompting strategies to further close the remaining optimality gap. Beyond open-pit mining, the simulator-in-the-loop architecture generalizes to any domain requiring long-horizon physical sequencing under hard operational constraints, advancing the broader goal of intelligent, interpretable automation in structured industrial systems.

\bibliographystyle{ACM-Reference-Format}
\bibliography{refe}

@INCOLLECTION{Altiti21,
  author={Altiti, Awwad H and Alrawashdeh, Rami O and Alnawafleh, Hani M},
  title     = {Open Pit Mining},
  booktitle = {Mining Techniques - Past, Present and Future},
  publisher = {IntechOpen},
  address   = {London},
  year      = {2021},
  doi       = {https://doi.org/10.5772/intechopen.92208}
}

@ARTICLE{cite2,
  title={Optimized open pit mine design, pushbacks and the gap problem—a review},
  author={Meagher, C and Dimitrakopoulos, R and Avis, D},
  journal={Journal of Mining Science},
  volume={50},
  number={3},
  pages={508--526},
  year={2014},
  publisher={Springer},
  doi = {https://doi.org/10.1134/S1062739114030132}
}

@ARTICLE{amponsah,
  author={Amponsah, Shadrach Yaw and Takouda, Pawoumodom Matthias and Ben-Awuah, Eugene},
  title   = {Genetic Algorithm Framework for Stochastic Open Pit Production Scheduling in the Presence of Grade Uncertainty},
  journal = {Mining Optimization Laboratory},
  volume  = {1},
  number  = {780},
  pages   = {95},
  year    = {2022}
}

@BOOK{johnson,
  author={Johnson, Thys Brentwood},
  title     = {Optimum Open Pit Mine Production Scheduling},
  publisher = {University of California, Berkeley},
  year      = {1968},
}

@ARTICLE{askari,
  title={Mixed integer linear programming formulations for open pit production scheduling},
  author={Askari-Nasab, Hooman and Pourrahimian, Yashar and Ben-Awuah, Eugene and Kalantari, Samira},
  journal={Journal of Mining Science},
  volume={47},
  number={3},
  pages={338--359},
  year={2011},
  publisher={Springer},
  doi = {https://doi.org/10.1134/S1062739147030117}
}

@TECHREPORT{Khazaei,
  author      = {Khazaei, Soroush and Pourrahimian, Yashar},
  title       = {A Comparative Analysis of Mathematical and Industrial Approaches for Sublevel Caving Long-Term Production Scheduling Optimization},
  institution = {Mining Optimization Laboratory, University of Alberta},
  address     = {Edmonton, Canada},
  number      = {Report Twelve, Paper 102},
  pages       = {48--76},
  year        = {2024}
 }

@ARTICLE{rahnema,
  author={Rahnema, Milad and Grenon, Martin and Moradi Afrapoli, Ali},
  title   = {Sustainable Open Pit Mining Through GHG-Conscious Short-Term Production Scheduling},
  journal = {International Journal of Mining, Reclamation and Environment},
  volume  = {39},
  number  = {5},
  pages   = {325--346},
  year    = {2025},
  doi = {https://doi.org/10.1080/17480930.2024.2394813}
}

@ARTICLE{Tabesh,
  author={Tabesh, Mohammad and Afrapoli, Ali Moradi and Askari-Nasab, Hooman},
  title   = {A Two-Stage Simultaneous Optimization of NPV and Throughput in Production Planning of Open Pit Mines},
  journal = {Resources Policy},
  volume  = {80},
  pages   = {103167},
  year    = {2023},
  doi = {https://doi.org/10.1016/j.resourpol.2022.103167}
}

@INPROCEEDINGS{hussain,
  author={Hussain, Sadam and Menon, Ramanunni Parakkal and Amara, Fatima and Lai, Chunyan and Eicker, Ursula},
  title     = {Optimization of Energy Systems using MILP and RC Modeling: A Real Case Study in Canada},
  booktitle = {IEEE International Systems Conference (SysCon)},
  pages     = {1--7},
  year      = {2024},
  doi = {https://doi.org/10.1109/SysCon61195.2024.10553577}
}

@ARTICLE{gptoss,
   title={gpt-oss-120b \& gpt-oss-20b model card},
  author={Agarwal, Sandhini and Ahmad, Lama and Ai, Jason and Altman, Sam and Applebaum, Andy and Arbus, Edwin and Arora, Rahul K and Bai, Yu and Baker, Bowen and Bao, Haiming and others},
  journal={arXiv preprint arXiv:2508.10925},
  year={2025},
  doi = {https://doi.org/10.48550/arXiv.2508.10925}
}

@MISC{ollama,
  author = {{Ollama}},
  title  = {gpt-oss on Ollama Library},
  year   = {2025},
  url    = {https://ollama.com/library/gpt-oss},
  note   = {[Accessed 25 January 2026]}
}

@ARTICLE{fontes2021analysis,
  author  = {Fontes, M. P. and Koppe, Jair Carlos and Silva Neto, J. A.},
  title   = {Analysis of the Variables: Commodity Price and Discount Rate on Long-Term Open Pit Mine Planning},
  journal = {Global Journal of Engineering and Technology Advances},
  volume  = {6},
  number  = {2},
  pages   = {142--150},
  year    = {2021},
  doi = {https://doi.org/10.30574/gjeta.2021.6.2.0025}
}

@MANUAL{gurobi,
  author = {Gurobi Optimization LLC},
  title  = {Gurobi Optimizer Reference Manual},
  year   = {2024},
  url    = {https://www.gurobi.com},
  note   = {[Accessed 18 December 2025]}
}

@ARTICLE{bubeck2023sparks,
  author={Bubeck, S{\'e}bastien and Chandrasekaran, Varun and Eldan, Ronen and Gehrke, Johannes and Horvitz, Eric and Kamar, Ece and Lee, Peter and Lee, Yin Tat and Li, Yuanzhi and Lundberg, Scott and others},
  title   = {Sparks of Artificial General Intelligence: Early Experiments with GPT-4},
  journal = {arXiv preprint arXiv:2303.12712},
  year    = {2023},
  doi = {https://doi.org/10.48550/arXiv.2303.12712}
}

@ARTICLE{da2025large,
title={Large Language Models for Combinatorial Optimization: A Systematic Review},
   ISSN={1557-7341},
   journal={ACM Computing Surveys},
   publisher={Association for Computing Machinery (ACM)},
   author={Da Ros, Francesca and Soprano, Michael and Di Gaspero, Luca and Roitero, Kevin},
   year={2026},
   doi ={https://doi.org/10.1145/3801961}
}

@INPROCEEDINGS{yang2023large,
title={Large Language Models as Optimizers},
author={Chengrun Yang and Xuezhi Wang and Yifeng Lu and Hanxiao Liu and Quoc V Le and Denny Zhou and Xinyun Chen},
booktitle={The Twelfth International Conference on Learning Representations},
year={2024}
}

@ARTICLE{app151810033,
  author={Zajac, Mateusz},
  title   = {Heuristic, Hybrid, and LLM-Assisted Heuristics for Container Yard Strategies Under Incomplete Information},
  journal = {Applied Sciences},
  volume  = {15},
  number  = {18},
  pages   = {10033},
  year    = {2025},
  doi={https://doi.org/10.3390/app151810033}
}

@ARTICLE{starjob2024,
 title={Starjob: Dataset for LLM-driven job shop scheduling},
  author={Abgaryan, Henrik and Cazenave, Tristan and Harutyunyan, Ararat},
  journal={arXiv preprint arXiv:2503.01877},
  year={2025},
  doi={https://doi.org/10.48550/arXiv.2503.01877}
}

@ARTICLE{afl2025,
title={An Agentic Framework with LLMs for Solving Complex Vehicle Routing Problems},
  author={Zhang, Ni and Cao, Zhiguang and Zhou, Jianan and Zhang, Cong and Ong, Yew-Soon},
  journal={arXiv preprint arXiv:2510.16701},
  year={2025},
  doi={https://doi.org/10.48550/arXiv.2510.16701}
}

@ARTICLE{lerchs1965,
  author  = {Lerchs, Helmut and Grossmann, Ingo F.},
  title   = {Optimum Design of Open-Pit Mines},
  journal = {CIM Bulletin},
  volume  = {58},
  number  = {633},
  pages   = {47--54},
  year    = {1965}
}

@inproceedings{
yao2023react,
title={ReAct: Synergizing Reasoning and Acting in Language Models},
author={Shunyu Yao and Jeffrey Zhao and Dian Yu and Nan Du and Izhak Shafran and Karthik R Narasimhan and Yuan Cao},
booktitle={The Eleventh International Conference on Learning Representations},
year={2023}
}

@INPROCEEDINGS{wei2022chain,
  author={Wei, Jason and Wang, Xuezhi and Schuurmans, Dale and Bosma, Maarten and Xia, Fei and Chi, Ed and Le, Quoc V and Zhou, Denny and others},
  title     = {Chain-of-Thought Prompting Elicits Reasoning in Large Language Models},
  booktitle = {Advances in Neural Information Processing Systems (NeurIPS)},
  volume    = {35},
  pages     = {24824--24837},
  year      = {2022}
}

@article{Dimitrakopoulos03072022,
author = {Roussos Dimitrakopoulos and Amina Lamghari},
title = {Simultaneous stochastic optimization of mining complexes - mineral value chains: an overview of concepts, examples and comparisons},
journal = {International Journal of Mining, Reclamation and Environment},
volume = {36},
number = {6},
pages = {443--460},
year = {2022},
publisher = {Taylor \& Francis},
doi = {https://doi.org/10.1080/17480930.2022.2065730}
}

@ARTICLE{jiang2025large,
  author={Jiang, Xia and Wu, Yaoxin and Li, Minshuo and Cao, Zhiguang and Zhang, Yingqian},
  title   = {Large Language Models as End-to-End Combinatorial Optimization Solvers},
  journal = {arXiv preprint arXiv:2509.16865},
  year    = {2025},
  doi = {https://doi.org/10.48550/arXiv.2509.16865}
}

@INPROCEEDINGS{loor2020applying,
  author={Loor, Valeria and Morales, Nelson},
  title     = {Applying Artificial Intelligence for Optimal Production Scheduling and Phase Design in Open Pit Mining},
  booktitle = {MassMin 2020},
  pages     = {1451--1466},
  year      = {2020},
  doi = {https://doi.org/10.36487/ACG_repo/2063_111}
}

@ARTICLE{saavedra2016optimizing,
  author={Saavedra-Rosas, Jos{\'e} and Jeivez, Enrique and Amaya, Jorge and Morales, Nelson},
  title   = {Optimizing Open-Pit Block Scheduling with Exposed Ore Reserve},
  journal = {Journal of the Southern African Institute of Mining and Metallurgy},
  volume  = {116},
  number  = {7},
  pages   = {655--662},
  year    = {2016},
  doi = {https://doi.org/10.17159/2411-9717/2016/v116n7a7}
}

@ARTICLE{caccetta2003application,
  author={Caccetta, Louis and Hill, Stephen P},
  title   = {An Application of Branch and Cut to Open Pit Mine Scheduling},
  journal = {Journal of Global Optimization},
  volume  = {27},
  number  = {2},
  pages   = {349--365},
  year    = {2003},
  doi = {https://doi.org/10.1023/A:1024835022186}
}

@ARTICLE{icarte2025intelligent,
  author={Icarte-Ahumada, Gabriel and Herzog, Otthein},
  title   = {Intelligent Scheduling in Open-Pit Mining: A Multi-Agent System with Reinforcement Learning},
  journal = {Machines},
  volume  = {13},
  number  = {5},
  pages   = {350},
  year    = {2025},
  doi = {https://doi.org/10.3390/machines13050350}
}

@ARTICLE{ieee,
  author={Yhdego, Tsegai O and Wang, Hui},
  title   = {Automated Ontology Generation for Zero-Shot Defect Identification in Manufacturing},
  journal = {IEEE Transactions on Automation Science and Engineering},
  year    = {2025},
  doi = {https://doi.org/10.1109/TASE.2025.3537463}
}

@article{UPADHYAY2018153,
title = {Simulation and optimization approach for uncertainty-based short-term planning in open pit mines},
journal = {International Journal of Mining Science and Technology},
volume = {28},
number = {2},
pages = {153-166},
year = {2018},
issn = {2095-2686},
doi = {https://doi.org/10.1016/j.ijmst.2017.12.003},
author = {Shiv Prakash Upadhyay and Hooman Askari-Nasab},
}

@article{KOUSHAVAND2014451,
title = {A linear programming model for long-term mine planning in the presence of grade uncertainty and a stockpile},
journal = {International Journal of Mining Science and Technology},
volume = {24},
number = {4},
pages = {451-459},
year = {2014},
issn = {2095-2686},
doi = {https://doi.org/10.1016/j.ijmst.2014.05.006},
author = {Behrang Koushavand and Hooman Askari-Nasab and Clayton V. Deutsch},
}

@article{LI2026207,
title = {A4PS: Agentic AI-assisted advanced planning and scheduling with large language models for smart manufacturing},
journal = {Journal of Manufacturing Systems},
volume = {85},
pages = {207-226},
year = {2026},
issn = {0278-6125},
doi = {https://doi.org/10.1016/j.jmsy.2026.01.003},
author = {Mingxing Li and Qu Zhou and Wanshan Li and Ting Qu and Maolin Yang and Pingyu Jiang}
}

@article{li2024pre,
  title={Pre-trained language models for text generation: A survey},
  author={Li, Junyi and Tang, Tianyi and Zhao, Wayne Xin and Nie, Jian-Yun and Wen, Ji-Rong},
  journal={ACM Computing Surveys},
  volume={56},
  number={9},
  pages={1--39},
  year={2024},
  publisher={ACM New York, NY}
}

@article{wang2025parameter,
  title={Parameter-efficient fine-tuning in large language models: a survey of methodologies},
  author={Wang, Luping and Chen, Sheng and Jiang, Linnan and Pan, Shu and Cai, Runze and Yang, Sen and Yang, Fei},
  journal={Artificial Intelligence Review},
  volume={58},
  number={8},
  pages={227},
  year={2025},
  publisher={Springer},
  doi = {https://doi.org/10.1007/s10462-025-11236-4}
}

@article{deepseekr1,
  title={Deepseek-r1: Incentivizing reasoning capability in llms via reinforcement learning},
  author={Guo, Daya and Yang, Dejian and Zhang, Haowei and Song, Junxiao and Wang, Peiyi and Zhu, Qihao and Xu, Runxin and Zhang, Ruoyu and Ma, Shirong and Bi, Xiao and others},
  journal={arXiv preprint arXiv:2501.12948},
  year={2025},
  doi = {https://doi.org/10.48550/arXiv.2501.12948}
}

@InProceedings{ma2024bilevel,
  title = 	 {{LLM} and Simulation as Bilevel Optimizers: A New Paradigm to Advance Physical Scientific Discovery},
  author =       {Ma, Pingchuan and Wang, Tsun-Hsuan and Guo, Minghao and Sun, Zhiqing and Tenenbaum, Joshua B. and Rus, Daniela and Gan, Chuang and Matusik, Wojciech},
  booktitle = 	 {Proceedings of the 41st International Conference on Machine Learning},
  pages = 	 {33940--33962},
  year = 	 {2024},
  editor = 	 {Salakhutdinov, Ruslan and Kolter, Zico and Heller, Katherine and Weller, Adrian and Oliver, Nuria and Scarlett, Jonathan and Berkenkamp, Felix},
  volume = 	 {235},
  series = 	 {Proceedings of Machine Learning Research},
  month = 	 {21--27 Jul},
  publisher =    {PMLR},
  url = 	 {https://proceedings.mlr.press/v235/ma24m.html},
}

@article{gao2024llm,
  title={Large language models empowered agent-based modeling and simulation: A survey and perspectives},
  author={Gao, Chen and Lan, Xiaochong and Li, Nian and Yuan, Yuan and Ding, Jingtao and Zhou, Zhilun and Xu, Fengli and Li, Yong},
  journal={Humanities and Social Sciences Communications},
  volume={11},
  number={1},
  pages={1--24},
  year={2024},
  publisher={Palgrave}
}

@article{maaef2025,
title = {Multi-agent large language models as evolutionary optimizers for scheduling optimization},
journal = {Computers \& Industrial Engineering},
volume = {206},
pages = {111197},
year = {2025},
issn = {0360-8352},
doi = {https://doi.org/10.1016/j.cie.2025.111197},
url = {https://www.sciencedirect.com/science/article/pii/S0360835225003432},
author = {Yidan Wang and Jiayin Wang and Zhiwei Chu}
}


\end{document}